\documentclass{article}
\usepackage{url}
\usepackage{graphicx}
\usepackage{amsfonts,amsmath,amssymb,amsthm}
\usepackage{bbm}
\usepackage{xr}
\usepackage{algorithmicx}
\usepackage{algpseudocode}
\usepackage{multicol}
\usepackage{natbib}
\newcommand{\R}{\ensuremath \mathbb{R}}

\newcommand{\T}{\ensuremath \mathbb{T}
}\newcommand{\Tcal}{\ensuremath \mathcal{T}}
\newcommand{\E}{\ensuremath \mathbf{E}}

\newcommand{\KL}{\ensuremath \mathrm{KL}}
\newcommand{\Mtilde}{\ensuremath \widetilde{M}}

\renewcommand{\b}{\ensuremath \mathbf{b}}
\newcommand{\W}{\ensuremath \mathbf{W}}
\renewcommand{\P}{\ensuremath \mathbf{P}}

\newcommand{\argmax}{\ensuremath \arg\max}

\newcommand{\indicator}[1]{\mathbbm{1}{\left[ {#1} \right] }}

\theoremstyle{plain}
\newtheorem{theorem}{Theorem}

\newtheorem{property}{Property}

\theoremstyle{definition}
\newtheorem{definition}{Definition}

\theoremstyle{remark}
\newtheorem{remark}{Remark}

\title{Correlated Equilibria for Approximate Variational Inference in MRFs}

\author{Luis E.~Ortiz, Boshen Wang \\Department of Computer \& Information Science\\
 University of Michigan - Dearborn\\
 \texttt{\{leortiz,boshenw\}@umich.edu} \and Ze Gong\\ School of Computing, Informatics \& Decision Systems  Engineering\\
 Arizona State University\\
 \texttt{Ze.Gong@asu.edu}}

\begin{document}

\maketitle

%

%






\begin{abstract}
Almost all of the work in graphical models for game theory has
mirrored previous work in probabilistic graphical models. Our work
considers the opposite direction: Taking advantage of recent advances
in equilibrium computation for probabilistic inference. In particular,
we present formulations of inference problems in Markov random
fields (MRFs) as computation of equilibria in a certain class of
game-theoretic graphical models. While some previous work explores
this direction, none of that work concretely establishes the precise
connection between variational 
probabilistic inference in MRFs and correlated
equilibria. There is no work that exploits recent theoretical and empirical
results from the literature on algorithmic and computational game theory on the tractable, polynomial-time computation of exact or
approximate correlated equilibria in
graphical games with arbitrary, loopy graph structure. Our work
discusses how to design new 
algorithms with equally tractable guarantees for the
computation of
approximate variational inference in MRFs. In addition, inspired by 
a previously stated game-theoretic view of state-of-the-art tree-reweighed (TRW) message-passing
techniques for belief inference as zero-sum game, we propose a different, general-sum potential game
to design approximate fictitious-play
techniques. We perform synthetic experiments evaluating our proposed
approximation algorithms with standard methods and TRW on several
classes of classical Ising models (i.e., with binary random variables). We also evaluate the algorithms using Ising models learned from the MNIST dataset.
Our experiments show that our global approach is
competitive, particularly shinning in a class of Ising models with
constant, ``highly attractive'' edge-weights, in which it is often better than
all other alternatives we evaluated. With a notable exception, our more local approach was
not as effective as our global approach or TRW. Yet, in fairness, almost all of the
alternatives are often no better than a simple baseline: estimate the
marginal probability to be $0.5$.
\end{abstract}

\section{Introduction}

Almost all of the work in graphical games has borrowed heavily from
analogies to probabilistic graphical models.
Yet, over-reliance on those analogies and previous standard approaches to exact inference might have led that approach to face the same computational roadblocks that plagued most exact-inference techniques. 

As an example of work that heavily exploits previous work in probablistic graphical models (PGMs),~\citet{kakade03} designed polynomial-time algorithms based on
linear programming for computing
 \emph{correlated equilibria (CE)} in standard graphical games with tree graphs.
The approach and polynomial-time results extend to
graphical games with bounded-tree-width graphs and 
graphical polymatrix games with tree graphs.
 Exact inference is tractable in PGMs whose graphs have bounded treewidth, but
intractable in general~\citep{cooper90,shimony94,istrail00}. 
In \citeyear{papadimitriou05}, \citeauthor{papadimitriou05} 
showed the intractability of computing the ``social-welfare'' optimum CE in arbitrary graphical games (see also \citet{papadimitriou08}). Everything seemed to point toward an eventual resignation that the approach of \citet{kakade03}, along with any other approach to the problem for that matter, had hit the ``bounded-treewidth-threshold wall.''

Yet, 
soon after,
\citet{papadimitriou05_ce} took a radically different approach to the
problem, and surprised the community with an efficient algorithm for
computing CE not only in graphical games, but also in almost all known compactly representable games. \citet{Jiang2015347} built upon~\citeauthor{papadimitriou05_ce}'s idea to provide what most people would consider an improved polynomial-time algorithm, because of the simplification of the CE that their algorithm outputs (see also~\citealp{DBLP:journals/sigecom/JiangL11}, for a summary).~\footnote{\citeauthor{papadimitriou05_ce}'s work has an interesting history, which \citet{Jiang2015347} nicely summarize. Some questions arose at the time about the technical soundness in the description of some steps in~\citeauthor{papadimitriou05_ce}'s algorithm. \citet{Jiang2015347} provided clarifications to those steps.}

An immediate question that arises from the algorithmic results just described is, what is so fundamentally different between the problem of exact inference in graphical models
and equilibrium computation that made this result possible in the context of graphical games? Of course, CE, probabilistic inference, and their variants are different problems, even within the same framework of graphical models. The question is, 
how different are they?  

It is well-known that {\em pure strategy Nash equilibrium (PSNE)\/} is inherently a classical/standard discrete {\em constraints satisfaction problem (CSP)}. 
It is also well-known that any CSP can be cast as a most-likely, or equivalently, a {\em maximum a posteriori (MAP)\/} assignment estimation problem in {\em Markov random fields (MRFs)}.~\footnote{Assuming a solution exists, of course; otherwise the resulting MRF is not well-defined.}
Through this connection, it is clear that there exists a MAP formulation of PSNE. But what about other, more general forms of equilibria?

We present here 
a formulation of the problem of equilibrium computation as a kind of
local conditions for different approximations to belief
inference. Similarly, we show how one can view some special games,
called \emph{graphical potential games}~\citep{ortiz15}, as defining an {\em equivalent\/} MRF whose ``locally optimal'' solutions correspond to {\em arbitrary\/} equilibria of the game.
Hence, \citeauthor{papadimitriou05_ce}'s result, and later that of \citeauthor{Jiang2015347}, open up the possibility that at least new classes of problems in probabilistic graphical models could be solved exactly and efficiently. The question is, which classes?

While we provide specific connections between the two fields that yield immediate theoretical and computational implications, we also provide practical alternatives that result from those connections. That is, the foundation of both \citeauthor{papadimitriou05_ce}'s  and~\citeauthor{Jiang2015347}'s algorithms is the \emph{ellipsoid method}, which is one approach that leads to the polynomial-time algorithm for linear programming. This approach, while provably efficient in theory, is often seen as less practical as other alternatives such as so-called \emph{interior-point methods}. This is in contrast to the simple linear programs that are possible for certain classes of graphical games~\citep{kakade03}. 
Are there simpler and practically effective variants of \citeauthor{papadimitriou05_ce}'s or \citeauthor{Jiang2015347}'s algorithms? While the last question is an important open question, we do not address it directly in this paper. Instead, We employ ideas from the literature of learning in games~\citep{fudenberg99}, particularly no-regret algorithms and fictitious play, to propose two specific instances of game-theoretic inspired, practical, and effective heuristics for belief inference in MRFs. One heuristic takes a local approach, and the other takes a global approach. We evaluate our proposed algorithms within the context of the most popular, standard, and state-of-art techniques from the literature in probabilistic graphical models.

This manuscript describes our work, which starts to address some of the questions above, and reports 
on our progress.

\subsection{Overview of the Paper}

Section~\ref{sec:prelim} provides preliminary material, introducing basic notation, terminology, and concepts from graphical models and game theory. 


Section~\ref{sec:infeq} is the main technical section of the paper. It shows reductions of different problems in belief inference in MRFs as computing equilibria in graphical potential games compactly represented as {\em Gibbs potential games}~\citep{ortiz15}. The reductions presented here vary in generality from {\em MAP assignment, marginals, and full-joint estimation} to {\em pure-strategy Nash equilibria (PSNE), mixed-strategy Nash equilibria (MSNE), and correlated equilibria (CE)}, respectively. 
We briefly discuss a connection between~\citeauthor{papadimitriou05_ce}'s algorithm, as well as~\citeauthor{jiang15}'s, and the work of~\citet{jaakkola97} on variational approximations  to the problem of probabilistic inference in MRFs via mean-field mixtures.
The paper also includes a discussion on the connections to previous work in computer vision on the problem of relaxation labeling, and work on game-theoretic approaches to (Bayesian) statistical estimation.
We then present an alternative approach based on a more global view of the problem, in contrast to the more local approach of the formulations mentioned above. More specifically, we formulate the inference problem using a two-player potential game, inspired by the work on \emph{tree reweighed (TRW) message-passing}~\citep{1459045}. We propose a special type of sequential, ``hybrid'' standard and stochastic fictitious play algorithm for belief inference. 

Section~\ref{sec:exp} reports on our experimental evaluation.
We compare our proposed algorithms to the popular, most commonly used, standard, and easily implementable approximation techniques in use today.

Section~\ref{sec:fw} discusses future work and suggests new opportunities for other potential research directions, beyond those already discussed in the main technical sections of the paper.

Section~\ref{sec:cont} concludes the paper with a summary of our contributions.



\section{Preliminaries}
\label{sec:prelim}

This section introduces basic notation and concepts in graphical models and game theory used throughout the paper. It also includes brief statements on current state-of-the-art mathematical and computational results in the area.

\paragraph{Basic Notation.} Denote by $x \equiv (x_1,x_2,\ldots,x_n)$ an $n$-dimensional vector and by $x_{-i} \equiv (x_1,\ldots,x_{i-1},x_{i+1},\ldots,x_n)$ the same vector without component $i$. Similarly, for every set $S \subset [n] \equiv \{1,\ldots,n\}$, denote by $x_S \equiv (x_i : i \in S)$ the (sub-)vector formed from $x$ using only components in $S$, such that, letting $S^c \equiv [n] - S$ denote the complement of $S$, we can denote $x \equiv (x_S,x_{S^c}) \equiv (x_i,x_{-i})$ for every $i$. If $A_1,\ldots,A_n$ are sets, denote by $A \equiv \times_{i \in [n]} A_i$, $A_{-i} \equiv \times_{j \in [n] - \{i\}} A_j$ and $A_S \equiv \times_{j \in S} A_j$.

\paragraph{Graph Terminology and Notation.} Let $G = (V,E)$ be an undirected graph,
with finite set of $n$ {\em vertices\/} or {\em nodes\/} $V = \{1, \ldots, n\}$ and a set of (undirected) edges $E$.
For each node $i$, let $\mathcal{N}(i) \equiv \{ j \mid (i,j) \in E \}$ be the set of neighbors of $i$ in $G$, {\em not including\/} $i$, and $N(i)  \equiv \mathcal{N}(i) \cup  \{i\}$ the set {\em including\/} $i$.
A {\em clique\/} $C$
of $G$ is a set of nodes with the property that they are all mutually connected: for all $i, j \in C$, $(i,j) \in E$; in addition, $C$ is {\em maximal\/} if there is no other node $k$ outside $C$ that is also connected to each node in $C$, i.e., for all $k \in V - C$, $(k,i) \notin E$ for some $i \in C$.

Another useful concept in the context of this paper is that of hypergraphs, which are generalizations of regular graphs. A {\em hypergraph graph\/} $\mathcal{G} = (V,\mathcal{E})$ is defined by a set of nodes $V$ and a set of {\em hyperedges\/} $\mathcal{E} \subset 2^V$. We can think of the hyperedges as cliques in a regular graph. Indeed, the {\em primal graph\/} of the hypergraph is the graph induced by the node set $V$ and where there is an edge between two nodes if they both belong to the same hyperedge; in other words, the primal graph is the graph induced by taking each hyperedge and forming cliques of nodes in a regular graph.


\subsection{Probabilistic Graphical Models}

Probabilistic graphical models are an elegant marriage of probability and graph theory that has had tremendous impact in the theory and practice of modern artificial intelligence, machine learning, and statistics. It has permitted effective modeling of large, structured high-dimensional complex systems found in the real world. The language of probabilistic graphical models allows us to capture the structure of complex interactions between individual entities in the system within a single model. The core component of the model is a graph in which each node $i$ corresponds to a random variable $X_i$ and the edges express conditional independence assumptions about those random variables in the probabilistic system. 

\subsubsection{Markov Random Fields, Gibbs Distributions, and \\ the Hammersley-Clifford Theorem} 
By definition, a joint probability distribution $P$ is a {\em Markov random field (MRF)\/} with respect to (wrt) an undirected graph $G$ if for all $x$, for every node $i$,
\(
P(X_i = x_i \mid X_{-i} = x_{-i}) = P(X_i = x_i \mid X_{\mathcal{N}(i)} = x_{\mathcal{N}(i)}).
\)
In that case, the neighbors/variables $X_{\mathcal{N}(i)}$ form the {\em Markov blanket\/} of node/variable $X_i$. 

Also by definition, a joint distribution $P$ is a {\em Gibbs distribution\/} wrt an undirected graph $G$ if 
it can be expressed as 
\(
\textstyle
P(X = x) = \prod_{C \in \mathcal{C}} \Phi_C(x_C)
\)
for some functions $\Phi_C$ indexed by a clique $C \in \mathcal{C}$, the set of all (maximal) cliques in $G$, and mapping every possible value $x_C$ that the random variables $X_c$ associated with the nodes in $C$ can take to a non-negative number. 

We say that a joint probability distribution $P$ is {\em positive\/} if it has full support (i.e., $P(x) > 0$ for all $x$).~\footnote{The positivity constraint is only necessary for the ``only if'' case proof of the theorem.}
\begin{theorem}
\label{thm:hc}
{\bf (Hammersley-Clifford~\citep{hammersley71})} Let $P$ be a positive joint probability distribution. Then, $P$ is an MRF with respect to $G$ if and only if $P$ is a Gibbs distribution with respect to $G$.
\end{theorem}

\noindent In the context of the theorem, the functions $\Phi_C$ are positive, which allows us to define MRFs in terms of {\em local potential functions\/} $\{ \phi_C \}$ over each clique $C$ in the graph. 
Define the function $\Psi(x) \equiv \sum_{C \in \mathcal{C}} \phi_C(x_C)$. Let us refer to any function of this form as a {\em Gibbs potential\/} with respect to $G$. A more familiar expression of an MRF is
\(
\textstyle
P(X = x) \propto \exp(\sum_{C \in \mathcal{C}} \phi_C(x_C)) = \exp(\Psi(x)) .
\)

\subsubsection{Some Inference-Related Problems in MRFs}

One problem of interest in an MRF is to compute a {\em most likely assignment\/} $x^* \in \argmax_x P(X = x) = \argmax_x \sum_{C \in \mathcal{C}} \phi_C(x_C))$; that is, the most likely outcome with respect to the MRF $P$. Another problem is to compute the {\em individual marginal probabilities\/} $P(X_i = x_i) = \sum_{x_{-i}} P(X_i = x_i, X_{-i} = x_{-i}) \propto \sum_{x_{-i}} \exp(\sum_{C \in \mathcal{C}} \phi_C(x_C)))$ for each variable $X_i$. A related problem is to compute the normalizing constant $Z = \sum_x \exp(\sum_{C \in \mathcal{C}} \phi_C(x_C)))$ (also known as the {\em partition function\/} of the MRF). 

Another set of problems concern so called ``belief updating.'' That is, computing information related to the {\em posterior probability distribution\/} $P'$ having observed the outcome of some of the variables, also known as the {\em evidence}. For MRFs, this problem is computationally equivalent to that of 
computing \emph{prior} marginal probabilities.

\subsubsection{Brief Overview of Computational Results in \\
Probabilistic Graphical Models} Both the exact and approximate versions of most inference-related problems in MRFs are in general intractable (e.g., NP-hard),
although polynomial-time algorithms do exists for some special cases
(see, e.g., \citealp{DAGUM1993141}, \citealp{ROTH1996273}, \citealp{istrail00}, \citealp{Wang20131610}, and the references therein).
The complexity of exact algorithms is usually characterized by structural properties of the graph, and the typical statement is that running times are polynomial only for graphs with bounded treewidth (see, e.g., \citealp{russell03} for more information). Several deterministic and randomized approximation approaches exist (see, e.g., \citealp{jordan99,jaakkola00,geman84}). 
An approximation approach of particular
interest in this paper is \emph{variational
  inference}~\citep{jordan99,jaakkola00}. Roughly speaking, the
general idea is to approximate an intractable MRF $P$ by a ``closest'' probability
distribution $Q^*$ within a ``computationally tractable'' class
$\mathcal{Q}$: formally, $Q^* \in \argmax_{Q \in \mathcal{Q}} \KL(Q
\parallel P)$, where $\KL(Q
\parallel P) \equiv \sum_x Q(x) \ln \frac{Q(x)}{P(x)}$ is the \emph{Kullback-Leibler (KL) divergence} between
probability distributions $P$ and $Q$ wrt $Q$. The
simplest example in the so called \emph{mean-field (MF)
  approximation}, in which $\mathcal{Q} = \{ Q \mid Q(x) = \prod_i
Q(x_i) \text{ for all } x \in \Omega \}$ consists of all possible
\emph{product} distributions. Even if $P$ is an IM, no closed-form
solution exists for its mean-field approximation, and the most common
computational scheme is based on simple axis parallel optimizations,
leading to individual local conditions of optimality and potential
local minima: that is, the problem is essentially reduced to finding
$Q^*(x) = \prod_i Q^*_i(x_i)$ such that for all $i$, we have
$Q^*_i \in
\argmax_{Q_i} \sum_{x_i} Q_i(x_i) \sum_{x_{-i}} \left[ \prod_{j \neq i}
Q^*_j(x_j) \right] \Psi(x_i,x_{-i}) + H_{Q_i}(X_i)$, where
$H(Q_i) \equiv H_{Q_i}(X_i) \equiv - \sum_{x_i} Q_i(x_i) \ln Q_i(x_i)$ is
the (Shannon) entropy of random variable $X_i \sim Q_i$.

\subsection{Game Theory}

Game theory~\citep{vonneumann47} provides a mathematical model of the stable behavior (or outcome) that may result from the interaction of rational individuals. 
This paper concentrates on {\em noncooperative\/} settings: individuals maximize their {\em own\/} utility, act  {\em independently}, and do not have (direct) control over the behavior of others.~\footnote{Individual rationality here 
means that each player seeks to maximize their own utility. Also note that, while many parlor ``win-lose''/zero-sum games involve competition, in general, {\em noncooperative $\neq$ competitive}: each player just wants to do the best for himself, regardless of how useful or harmful his behavior is to others.
}  

The concept of {\em equilibrium\/} is central to game theory. 
Roughly, an equilibrium in a noncooperative game is a point of strategic stance, where no individual player can gain by {\em unilaterally\/} deviating from the equilibrium behavior.

\subsubsection{Games and their Representation} 
\label{sec:game_rep}
Let $V = [n]$
denote a finite set of $n$ players in a game. For each player $i \in V$, let $A_i$ denote the set of {\em actions} or {\em pure strategies\/} that $i$ can play. Let $A \equiv \times_{i \in V} A_i$ denote the set of {\em joint actions\/}, $x \equiv (x_i, \ldots, x_n) \in A$ denote a joint action, and $x_i$ the individual action of player $i$ in $x$. Denote by $x_{-i} \equiv (x_1, \ldots, x_{i-1}, x_{i+1}, \ldots, x_n)$ the joint action of all the players {\em except\/} $i$, such that $x \equiv (x_i, x_{-i})$. Let $M_i : A \to \R$ denote the {\em payoff/utility function\/} of player $i$. If the $A_i$'s are finite, 
then $M_i$ is called the {\em payoff  matrix} of player $i$. Games represented this way are called {\em normal-\/} or {\em strategic-form games}.

There are a variety of compact representations for large games inspired by probabilistic graphical models in AI and machine learning~\citep{la-mura00,kearns01,koller03,leyton-brown03,jiang08}. The results of this paper are presented in the context of the following generalization of {\em graphical games}~\citep{kearns01}, a simple but powerful model inspired by probabilistic graphical models such as MRFs previously defined by~\citet{ortiz14}.~\footnote{Connections have already been established between the different kinds of compact representations \citep{jiang08}, which may facilitate extensions of ideas, frameworks, and results to those alternative models.}
 
\begin{definition}
A {\em graphical multi-hypermatrix game (GMhG)\/} is defined by
\begin{itemize}
\item a {\em directed\/} graph $G = (V,E)$ in which there is a node $i \in V$ in $G$ for each of the $n$ players in the game (i.e., $|V| = n$), and the set of directed edges, or arcs, $E$ defines a set of {\em neighbors\/} $\mathcal{N}(i) \equiv \{ j \mid  (j,i) \in E, i \neq j \}$ whose action affect the payoff function of $i$ (i.e., $j$ is a neighbor of $i$ if and only if there is an arc from $j$ to $i$); and
\item for each player $i \in V$, 
\begin{itemize}
\item a set of actions $A_i$,
\item a hypergraph where the vertex set is its {\em (inclusive) neighborhood\/} $N(i) \equiv \mathcal{N}(i) \cup \{ i \}$ and the hyperedge set is a set of {\em cliques\/} of players $\mathcal{C}_i \subset 2^{N(i)}$, and
\item a set $\{M'_{i,C} : A_C \to \R  \mid C \in \mathcal{C}_i \}$ of {\em local-clique payoff (hyper)matrices\/}. 
\end{itemize}
\end{itemize}
The interpretation of a GMhG is that, for each player $i$, the {\em local\/} and {\em global payoff (hyper)matrices\/} $M'_i : A_{N(i)} \to \R$ and $M_i : A \to \R$ of $i$ are (implicitly) defined as $M'_i(x_{N(i)}) \equiv \sum_{C \in \mathcal{C}_i} M'_{i,C}(x_{C})$ and $M_i(x) \equiv M'_i(x_{N(i)})$, respectively. 
\end{definition}

\paragraph{Graphical potential games.} Graphical potential games are
special instances of GMhGs. They play a key role in establishing a stronger connection between probabilistic inference in MRFs and equilibria in games than previously noted. 
\citet{ortiz15} provides a characterization of graphical potential games, and discusses the implication of convergence of certain kinds of ``playing'' processes in games based on connections to the Gibbs sampler~\citep{geman84}, via the Hammersley-Clifford Theorem~\citep{hammersley71,besag74}. 
\citet{yu95} (implicitly) used graphical potential games to establish an equivalence between {\em local\/} maximum-{\em a-posteriori\/} (MAP) inference in Markov random fields and Nash equilibria of the game, a topic revisited in Section~\ref{sec:pneinf}.~\footnote{In the interest of brevity, please see~\citet{ortiz14} for a thorough discussion of GMhGs, including their compact representation size and connections to other classical classes of games in game theory.}

\subsubsection{Equilibria as Solution Concepts} Equilibria are generally considered {\em the\/} solutions of games. Various notions of equilibria exist. 
A {\em pure strategy (Nash) equilibrium (PSNE)\/} of a game is a joint action $x^*$ such that for all players $i$, and for all actions $x_i$,
\(
M_i(x^*_i, x^*_{-i}) \geq M_i(x_i, x^*_{-i}).
\)
That is, no player can improve its payoff by {\em unilaterally\/} deviating from its prescribed equilibrium $x^*_i$, assuming the others stick to their actions $x^*_{-i}$. Some games, such as the extensively-studied Prisoner's Dilemma, have PSNE;
many others, such as ``playground'' Rock-Paper-Scissors, do not. 
This is problematic 
because
it will not be possible to ``solve'' some games using PSNE. 

A {\em mixed-strategy\/} of player $i$ is a probability distribution $Q_i$ over $A_i$ such that $Q(x_i)$ is the probability that $i$ chooses to play action $x_i$.~\footnote{Note that the sets of mixed strategies contain pure strategies, as we can always recover playing a pure strategy exclusively.}  A {\em joint mixed-strategy\/} is a joint probability distribution $Q$ capturing the players behavior, such that $Q(x)$ is the probability that joint action $x$ is played, or in other words, each player $i$ plays action in component $x_i$ of $x$. Because we are assuming that the players play {\em independently}, $Q$ is a product distribution: $Q(x) = \prod_{i} Q_i(x_i)$. 
Denote by $Q_{-i}(x_{-i}) \equiv \prod_{j \neq i} Q_j(x_j)$ the joint mixed strategies of all the players except $i$. The {\em expected payoff\/} of a player $i$ when some joint mixed-strategy $Q$ is played is $\sum_x Q(x) M_i(x)$; abusing notation, denote it by $M_i(Q)$. The {\em conditional expected payoff\/} of a player $i$ given that he plays action $x_i$ is $\sum_{x_{-i}} Q_{-i}(x_{-i}) M_i(x_i, x_{-i})$; abusing notation again, denote it by $M_i(x_i,Q_{-i})$.

A {\em mixed-strategy Nash equilibrium (MSNE)\/} is a joint mixed-strategy $Q^*$ that is a product distribution formed by the individual players mixed strategies $Q^*_i$ such that, for all players $i$, and any other alternative mixed strategy $Q'_i$ for his play,
\(
M_i(Q^*_i, Q^*_{-i}) \geq M_i(Q'_i, Q^*_{-i}).
\)
{\em Every\/} game in normal-form has at least one such equilibrium~\citep{nash51}. 
Thus, every game has an MSNE ``solution.''

One relaxation of MSNE considers the case where the amount of gain each player can obtain from unilateral deviation is very small. This concept is particularly useful to study approximation versions of the computational problem.  Given $\epsilon \geq 0$, an {\em (approximate) $\epsilon$-Nash equilibrium (MSNE)\/} is defined as above, except that the expected gain condition becomes
\(
M_i(Q^*_i, Q^*_{-i}) \geq M_i(Q'_i, Q^*_{-i}) - \epsilon.
\)

Several refinements and generalizations of MSNE have been proposed. One of the most interesting generalizations is that of a {\em correlated equilibrium (CE)}
 \citep{aumann74}. In contrast to MSNE, a CE can be a full joint distribution, and thus characterize more complex joint-action behavior by players. 
Formally, a {\em correlated equilibrium (CE)\/} is a joint probability distribution $Q$ over $A$ such that, for all players $i$, $x_i, x_i' \in A_i$, $x_i \neq x'_i$, and $Q(x_i) > 0$,
\[
\sum_{x_{-i}} Q(x_{-i} | x_i) M_i(x_i, x_{-i}) \geq \sum_{x_{-i}} Q(x_{-i}|x_i) M_i(x'_i, x_{-i}),
\]
where $Q(x_i) \equiv \sum_{x_{-i}} Q(x_i,x_{-i})$ is the (marginal) probability that player $i$ will play $x_i$ according to $Q$ and $Q(x_{-i} | x_i) \equiv Q(x_i,x_{-i}) / \sum_{x'_i} Q(x'_i,x_{-i})$ is the conditional given $x_i$. 
An MSNE is CE that is a product distribution.
An equivalent expression of the CE condition above
is
\(
\textstyle
\sum_{x_{-i}} Q(x_i, x_{-i}) M_i(x_i, x_{-i}) \geq \sum_{x_{-i}} Q(x_i, x_{-i}) M_i(x'_i, x_{-i}).
\)
As was the case for MSNE, we can relax the condition of deviation to account for potential gains from small deviation. Given $\epsilon > 0$, adding the term ``$-\epsilon$'' to the right-hand-side of the condition above defines an {\em (approximate) $\epsilon$-CE\/}.~\footnote{Note that approximate CE is usually defined based on this unconditional version of the CE conditions~\citep{hart00}.}

CE have several conceptual and computational advantages over MSNE. For instance, all players may achieve better expected payoffs in a CE than those achievable in any MSNE;~\footnote{The distinction between installing a traffic light at an intersection and leaving the intersection without one is a real-world example of this.} some ``natural'' forms of play are guaranteed to converge to the (\emph{set} of) CE~\citep{foster97,foster99,fudenberg99,hart00,hart03,hart05}; and CE is consistent with a Bayesian framework~\citep{aumann87}, something not yet possible, and apparently unlikely for MSNE~\citep{hart07}.

\subsubsection{Brief Overview of Results in Computational Game Theory} 
There has been an explosion of computational results on different equilibrium concepts on a variety of game representations and settings since the beginning of this century. The following is a brief summary. We refer the reader to a book by~\citet{nisan07} for a (partial) introduction to this 
research area.

The problem for two-player {\em zero-sum\/} games, where the sum of
the entries of both matrix is zero, and therefore only one matrix is
needed to represent the game, can be solved in polynomial time: It is
equivalent to linear
programming~\citep{vonneumann47,szep85,karlin59}. After being open for
over 50 years, the problems of the complexity of computing MSNE in
games was finally settled recently, following a very rapid sequence of
results in the last part of
2005~\citep{goldberg05_v2,daskalakis05,daskalakis05_three,daskalakis09_acm,chen05_tr_three}:
Computing MSNE is likely to be hard in the worst case, i.e.,
PPAD-complete~\citep{papadimitriou94}, even in games with only two
players~\citep{chen05_tr,chen06,chen09,daskalakis09,daskalakis09_acm}. The
result of \citet{fabrikant04} suggests that computing PSNE in
succinctly representable games is also likely to be intractable in the
worst case, i.e., PLS-complete~\citep{johnson88}. A common statement
is that computing MSNE, and in some cases even PSNE, with ``special
properties'' is hard in the worst
case~\citep{gilboa89,gottlob03,conitzer08}. Computing approximate MSNE
is also thought to be hard in the worst
case~\citep{chen06_approx,chen09}. We refer the reader
to~\citet{ortiz_and_irfan_17}, and the references therein, for recent
results along this line and a brief survey of the state-of-the-art for this problem.

Most current results for computing exact and approximate PSNE or MSNE in graphical games essentially mirror those for MRFs and constraint networks: polynomial time for bounded treewidth graph; intractable in general~\citep{kearns01,gottlob03,daskalakis06_mrf,ortiz14}. This is unsurprising because they were mostly inspired by analogous versions in probabilistic graphical models and constraint networks in AI, and therefore share similar characteristics.  Several heuristics exist for dealing with general graphs~\citep{vickrey02,ortizandkearns03,daskalakis06_mrf}. 

In contrast, there exist polynomial-time algorithms for computing CE, both for normal-form games (where the problem reduces to a simple linear feasibility problem) and
even most succinctly-representable games known today~\citep{papadimitriou05_ce,Jiang2015347}, including graphical games.

\section{Equilibria and Inference}
\label{sec:infeq}

The line of work presented in this section is partly motivated by the following question: {\em Can we leverage advances in computational game theory for problems in the probabilistic graphical models community?} Establishing a strong bilateral connection between both problems may help us answer this question.

The literature on computing equilibria in games has skyrocketed since the beginning of this century. 
As we discover techniques developed early on within the game theory community, and as new results are generated from the extremely active computational game theory community, we may be able to adapt those techniques for solving games to the inference setting. If we can establish a strong bilateral connection between inference problems and the computation of equilibria, we may be able to relate algorithms in both areas and exchange previously unknown results in each. 

\subsection{Pure-Strategy Nash Equilibrium and \\Approximate MAP Inference}
\label{sec:pneinf}

Consider an MRF $P$ with respect to graph $G$ and Gibbs potential $\Psi$ defined by the set of potential functions $\{ \phi_C \}$. For each node $i$, denote by $\mathcal{C}_i \subset \mathcal{C}$ the subset of cliques in $G$ that include $i$. Note that the (inclusive) neighborhood of player $i$ is given by $N(i) = \cup_{C \in \mathcal{C}_i} C$.

Define an {\em MRF-induced\/} GMhG, and more specifically, a (hyperedge-symmetric) hypergraphical game~\citep{papadimitriou05_ce,ortiz15}, with the same graph $G$, and for each player $i$, hypergraph with hyperedges $\mathcal{C}_i$ and local-clique payoff hypermatrices $M'_{i,C}(x_C) \equiv \phi_C(x_C)$ for all $C \in \mathcal{C}_i$. A few observations about the game are in order. 
\begin{property} 
The representation size of the MRF-induced game is the same as that of the MRF: not exponential in the largest neighborhood size, but the size of the largest clique in $G$.
\end{property}
\begin{property}
The MRF-induced game is a graphical potential game~\citep{ortiz15}
with graph $G$ and (Gibbs) potential function $\Psi$: i.e., for all
$i$, $x$ and $x'_i$, $M_i(x_i, x_{-i}) - M_i(x'_i, x_{-i}) = M'_i(x_i, x_{\mathcal{N}(i)}) - M'_i(x'_i, x_{\mathcal{N}(i)})$
\begin{align*}
= &\sum_{C \in \mathcal{C}_i} \phi_C(x_i, x_{C - \{i\}}) - \sum_{C \in \mathcal{C}_i} \phi_C(x'_i, x_{C - \{i\}}) \\
= & \sum_{C \in \mathcal{C}_i} \phi_C(x_i, x_{C - \{i\}}) + \sum_{C' \in \mathcal{C} - \mathcal{C}_i} \phi_{C'}(x_{C'}) + \\
 & - \sum_{C \in \mathcal{C}_i} \phi_C(x'_i, x_{C - \{i\}}) - \sum_{C'
   \in \mathcal{C} - \mathcal{C}_i} \phi_{C'}(x_{C'}) \\
 = & \Psi(x_i, x_{-i}) - \Psi(x'_i, x_{-i}).
\end{align*}
\end{property}


\begin{remark}
Through the connection established by the last property, it is easy to see that {\em sequential\/} best-response dynamics is guaranteed to converge to a PSNE of the game in finite time, regardless of the initial play.~\footnote{ Recall that {\em  best-response dynamics\/} refers to the a process where at each time step, each player observes the action $x_{-i}$ of others and takes an action that maximizes its payoff given that the others played $x_{-i}$. In this case, those dynamics would essentially be implementing an axis-parallel coordinate maximization over the space of assignments for the MRF, which is guaranteed to converge to a local maxima (or critical points) of the MRF.} In fact, we can conclude that a joint-action $x^*$ is a PSNE of the game if and only if $x^*$ is a local maxima or a critical point of the MRF $P$. Thus, the MRF-induced game, like {\em all\/} potential games~\citep{Monderer_and_Shapley_1996}, always has PSNE.~\footnote{This result should not be surprising given that other researchers have established a one-to-one relationship between the complexity class PLS~\citep{johnson88}, which characterizes local search problems, of which finding local maxima of the MRF is an instance, and (ordinal) potential games~\citep{fabrikant04}. } 

Similarly, for any potential game, one can define a {\em game-induced MRF\/} using the potential function of the game whose set of local maxima (and critical points) corresponds exactly to the set of PSNE of the potential game. Through this connection we can show that solving the local-MAP problem in MRFs is PLS-complete in general~\citep{fabrikant04}.~\footnote{A direct proof of this result follows from~\citet{papadimitriou90}, and in particular, the result for Hopfield neural networks~\citep{hopfield82}. A Hopfield neural network can be seen as an MRF, and more specifically, and Ising model, when the weights on the edges are symmetric. Similarly, any Hopfield neural network can be seen as a polymatrix game~\citep{miller92}; when the weights are symmetric the network can be seen as a potential game (in particular, it is an instance of a {\em party affiliation game}~\citep{fabrikant04}). Indeed, a stable configuration in an arbitrary Hopfield neural network is equivalent to a PSNE of a corresponding polymatrix game.  (See \citealp{papadimitriou90}, and \citealp{miller92}, for the relevant references.)}

One question that comes to mind is whether one can say anything about the properties of the globally optimal assignment in the game-induced MRF and the payoff it supports for the players. Or whether it can be characterized by stronger notions of equilibria. For example, are {\em strong NE}, in which no {\em coalition\/} of players could obtain a Pareto dominated set of payoffs by unilaterally deviating, joint MAP assignments of the MFR? Or more generally, what characteristics can we assign to the MAP assignments of the game-induced MRF?

In short, we can use algorithms for PSNE as heuristics to compute locally optimal MAP assignments of $P$ and {\em vice versa}.~\footnote{Note that algorithms for PSNE can in principle find critical points of $P$. In either case, algorithms such as the max-product version of belief propagation (BP) can only provide such local-optimum/critical-point convergence guarantees in general.}
\end{remark}

\begin{remark}
\citet{daskalakis07_random} extended results in game theory characterizing the number of PSNE in normal-form games (see \citealp{stanford95,rinott00}, and the references therein) to graphical games, but now taking into consideration the network structure of the game. Information about the number of PSNE in games can provide additional insight on the structure of MRFs.

For example, one of the results of~\citet{daskalakis07_random} states that for graphs respecting certain expansion properties as the number of nodes/players increases, the number of PSNE of the graphical game will have a limiting distribution that is a Poisson with expected value $1$. Also according to~\citet{daskalakis07_random}, a similar behavior occurs for games with graphs generated according to the Erd\"{o}s-R\'{e}nyi model with sufficiently high average-degree (i.e., reasonably high connectivity). Thus, either the set of MRF-induced games has significantly low measure relative to the set of all possible randomly generated games (something that seems likely), or the number of local maxima (and critical points) of the MRF will have a similar distribution, and thus that number is expected to be low. The latter would suggest that local algorithms such as the max-product algorithm may be less likely to get stuck in local maxima (or critical points) of the MRF.

In addition, there have been several results stating that PSNE are unlikely to exist in many graphs, and that, when they do exist, they are not that many~\citep{daskalakis07_random}.~\footnote{In particular, the number of PSNE has a Poisson distribution with parameter $1$.} MRF-induced games would in that sense represent a very rich class of {\em non-randomly generated\/} graphical games for which the results above do not hold. 

\end{remark}

\subsection{Mixed-strategy Equilibria and Belief Inference}

Going beyond PSNE and MAP estimation, this subsection begins to establish a stronger, and potentially more useful connection between probabilistic inference and more general concepts of equilibria in games. 


Let $S$
be a subset of the players (i.e., nodes in the graph) and denote by $Q_S(x_S) \equiv \sum_{x_{V-S}} Q(x)$ the (marginal) probability distribution of $Q$ over possible joint actions of players in $S$. Consider the condition for correlated equilibria (CE), which for the MRF-induced game we can express as, for all $i,x_i,x_i'\neq x_i$,
\begin{align*}
\sum_{x_{\mathcal{N}(i)}} Q_{N(i)}(x_i, x_{\mathcal{N}(i)}) \sum_{C
  \in \mathcal{C}_i} \phi_C(x_i, x_{C - \{i\}}) \geq \\ \sum_{x_{\mathcal{N}(i)}}  Q_{N(i)}(x_i,x_{\mathcal{N}(i)}) \sum_{C \in \mathcal{C}_i} \phi_C(x'_i, x_{C - \{i\}}) .
\end{align*}
Commuting the sums and simplifying we get the following equivalent condition:
\begin{align}
\textstyle
\nonumber & \sum_{C \in \mathcal{C}_i} \sum_{x_{C-\{i\}}} Q(x_i, x_{C - \{i\}})
\phi_C(x_i, x_{C - \{i\}}) \geq \\ 
\label{eqn:gibbsce}
&\sum_{C \in \mathcal{C}_i} \sum_{x_{C-\{i\}}}  Q(x_i,x_{C - \{i\}}) \phi_C(x'_i, x_{C - \{i\}}) .
\end{align}
This simplification is important because it highlights that, modulo expected payoff equivalence, we only need distributions over the original cliques, {\em not\/} the induced neighbohoods/Markov blankets, to represent CE in this class of games, in contrast to \citet{kakade03}; thus, we are able to maintain the size of the representation of the CE to be the same as that of the game. 

As an alternative, we can use the fact that the MRF-induced game is a potential game and, via some definitions and algebraic manipulation, get the following sequence of equivalent conditions, which hold for all $i$, $x_i$ and $x_i'$.
\begin{align*}
\textstyle
\sum_{x_{-i}} Q(x_i, x_{-i}) \left( M_i(x_i, x_{-i}) - M_i(x'_i, x_{-i}) \right) &\geq 0\\
\textstyle
\sum_{x_{-i}} Q(x_i, x_{-i}) \left( \Psi(x_i, x_{-i}) - \Psi(x'_i, x_{-i}) \right) & \geq 0 \\
\textstyle
\sum_{x_{-i}} Q(x_i, x_{-i}) \left( \ln P(x_i, x_{-i}) - \ln P(x'_i, x_{-i}) \right) & \geq 0
\end{align*}
Rewriting the last expression, we get the following equivalent condition: for all $i$, $x_i$ and $x_i'$,
\begin{align}
\label{eqn:cexent} 
\sum_{x_{-i}} Q(x_i, x_{-i}) [- \ln P(x_i, x_{-i})] \leq 
\textstyle
\sum_{x_{-i}} Q(x_i, x_{-i}) [- \ln P(x'_i, x_{-i})] \, .
\end{align}

The following are some additional remarks on the implications of the last condition.~\footnote{In what follows, we refer to concepts from information theory in the discussion, such as (Shannon's) entropy, cross entropy, and relative entropy (also known as Kullback-Leibler divergence). We refer the reader to~\citet{coverandthomas06} for a textbook introduction to those concepts.}

\begin{remark}
First, it is useful to introduce the following notation. For any distribution $Q'$, let $H(Q',P) \equiv \sum_{x} Q'(x) [- \log_2 P(x)]$ be the {\em cross entropy\/} between probability distributions $Q'$ and $P$, with respect to $P$.~\footnote{That is, (a lower bound on) the average number of bits required to transmit "messages/events" generated according to $Q$ but encoded using a scheme based on $P$.} Denote by $Q_{-i}(x_{-i}) \equiv \sum_{x_i} Q(x_i,x_{-i})$ the marginal distribution of play over the joint-actions of all players {\em except\/} player $i$. Denote by $Q'_i Q_{-i}$ the joint distribution defined as $(Q'_i Q_{-i}) (x) \equiv Q'_i(x_i) Q_{-i}(x_{-i})$ for all $x$.

Then, condition~\ref{eqn:cexent} implies the following sequence of conditions, which hold for all $i$.
\begin{align*}
\sum_{x} Q(x) [- \ln P(x)] & \leq 
\sum_{x_{-i}} Q_{-i}(x_{-i}) [- \ln P(x'_i, x_{-i})] \text{ for all } x'_i \\
H(Q,P) & \leq
\min_{x'_i}\sum_{x_{-i}} Q_{-i}(x_{-i}) [- \log_2 P(x'_i, x_{-i})] \\
& = 
\min_{Q'_i}\sum_{x} Q'_i(x_i) Q_{-i}(x_{-i}) [- \log_2 P(x_i, x_{-i})] \\
& = 
\min_{Q'_i}H(Q'_i Q_{-i}, P)
\end{align*}
As anonymous reviewer pointed out, the condition is actually that of a
\emph{coarse CE (CCE)}~\citep{hannan1957,Moulin1978}, which is a
superset of CE and allows us to apply several simple methods for computing such
equilibrium concept, as discussed later in this section.
Hence, {\em any CE of the MRF-induced game is a kind of approximate local optimum (or critical point) of an approximation of the MRF based on a special type of cross entropy minimization.} 

The following property summarizes this remark.
\end{remark}

\begin{property}
For any MRF $P$, any correlated equilibria $Q$ of the game induced by $P$ satisfies $H(Q,P) \leq \min_i \min_{Q'_i}H(Q'_i Q_{-i}, P)$.
\end{property}

\begin{remark}
Let us introduce some additional notation. For any joint distribution of play $Q'$, let $H(Q') \equiv \sum_x Q'(x) [-\log_2 Q'(x)]$ be its entropy. Similarly, for any player $i$, for any marginal/individual distribution of play $Q'_i$, let $H(Q'_i) \equiv \sum_{x_i} Q'_i(x_i) [-\log_2 Q'_i(x_i)]$ be its (marginal) entropy. For any distribution $Q'$ and $P$, let $\KL(Q' \parallel P) \equiv \sum_x Q'(x) \log_2 (Q'(x)/P(x)) = H(Q',P) - H(Q')$ be the {\em Kullback-Leibler divergence\/} between $Q'$ and $P$, with respect to $Q'$. Denote by $H(Q_{i \mid -i}) \equiv \sum_{x_i,x_{-i}} Q(x_i,x_{-i}) \log_2 (Q(x_i,x_{-i})/Q_{-i}(x_i)) = H(Q_{-i}) - H(Q)$ the conditional entropy of the individual play of player $i$ given the joint play of all the players except $i$, with respect to $Q$. 

Then, we can express the condition~\ref{eqn:cexent} as the following equivalent conditions, which hold for all $i$.
\begin{align*}
\textstyle
\KL(Q \parallel P) + H(Q) & \leq 
\min_{Q'_i} \KL(Q'_i Q_{-i} \parallel P) + H(Q'_i Q_{-i}) \\
\KL(Q \parallel P) + H(Q_{i \mid -i}) &
\leq
\min_{Q'_i} \KL(Q'_i Q_{-i} \parallel P) + H(Q'_i)
\end{align*}
Hence, {\em any CE of a MRF-induced game is a kind of approximate local optimum (or critical point) of a special kind of variational approximation of the MRF.}
The following property summarizes this remark.
\end{remark}

\begin{property}
\label{prop:cevar}
For any MRF $P$, any correlated equilibria $Q$ of the game induced by $P$ satisfies $\KL(Q \parallel P) \leq \min_i \left[ \min_{Q'_i} \KL(Q'_i Q_{-i} \parallel P) + H(Q'_i) \right] - H(Q_{i \mid -i}) $.
\end{property}
Note that the last property implies that the approximation $Q$ satisfies the local condition $\KL(Q \parallel P) \leq \min_i \min_{Q'_i} \KL(Q'_i Q_{-i} \parallel P) + \log_2 |\Omega_i|$.

Before continuing exploring connections to CE, it is instructive to first consider MSNE.

\subsubsection{Mixed-strategy Nash Equilibria and\\
 Mean-Field Approximations} In the special case of MSNE, the joint mixed strategy $Q(x) = \prod_i Q_i(x_i)$ is a product distribution. Denote by $Q^{\times}_{-i}(x_{-i}) \equiv \prod_{j \neq i} Q_j(x_j) = \sum_{x_i} Q(x)$ the (marginal) joint action of play over all the players except $i$, and denote by $(Q'_i Q^{\times}_{-i})$ the probability distribution defined such that the probability of $x$ is $(Q'_i Q^{\times}_{-i}) (x) \equiv Q'_i(x_i) Q^{\times}_{-i}(x_{-i})$. 

In this special case, the equilibrium conditions imply the following conditions, which hold for all $i$: for all $x_i$ such that $Q_i(x_i) > 0$,
\begin{align*}
& \sum_{x_{-i}} Q_i(x_i) Q^{\times}_{-i}(x_{-i}) [-\ln P(x_i,x_{-i})]
  \\ = & \min_{x'_i} \sum_{x_{-i}} Q_i(x_i) Q^{\times}_{-i}(x_{-i}) [-\ln P(x'_i,x_{-i})] \; .
\end{align*}
Denoting by $\mathcal{X}_i^+ \equiv \{ x_i \in A_i \mid Q_i(x_i) > 0
\}$, the last condition implies that
\begin{multline*}
\sum_{x_i \in \mathcal{X}_i^+} \sum_{x_{-i}} Q_i(x_i)
  Q^{\times}_{-i}(x_{-i}) [-\ln P(x_i,x_{-i})] = \\ \left( \sum_{x_i \in \mathcal{X}_i^+} Q_i(x_i) \right) \min_{x'_i} \sum_{x_{-i}} Q^{\times}_{-i}(x_{-i}) [-\ln P(x'_i,x_{-i})] \; .
\end{multline*}
The last condition is equivalent to
\begin{align*}
& \sum_{x_i} \sum_{x_{-i}} Q_i(x_i) Q^{\times}_{-i}(x_{-i}) [-\ln
  P(x_i,x_{-i})] \\ = & \min_{x'_i} \sum_{x_{-i}} Q^{\times}_{-i}(x_{-i}) [-\ln P(x'_i,x_{-i})] \; ,
\end{align*}
which, in turn, we can express as
\(
H(Q,P) = \min_{Q'_i} H(Q' Q^{\times}, P) \, .
\)
The last expression is also equivalent to
\[
\KL(Q \parallel P) + H(Q_i) =
\min_{Q'_i} \KL(Q'_i Q^{\times}_{-i} \parallel P) + H(Q'_i) \; . 
\]
Hence, {\em a NE $Q$ of the game is {\em almost\/} a locally optimal mean-field approximation, except for the extra entropic term.} In summary, for MSNE we have the following tighter condition than for arbitrary CE.

\begin{property}
For any MRF $P$, any MSNE $Q$ of the game induced by $P$ satisfies $\KL(Q \parallel P) = \left[ \min_{Q'_i} \KL(Q'_i Q^{\times}_{-i} \parallel P) + H(Q'_i) \right] - H(Q_i) $, for all $i$.
\end{property}
Note that the last property implies that the mean-field approximation $Q$ satisfies the local condition $\KL(Q \parallel P) \leq \min_{Q'_i} \KL(Q'_i Q^{\times}_{-i} \parallel P) + \log_2 |\Omega_i|$ for all $i$.

One possible way to address the issue of the extra entropic term is to
consider instead the {\em MRF-induced infinite game\/}, where each
player $i$ has the (continuous) utility function~\footnote{In an {\em
    infinite game\/} the sets of actions or pure strategies are
  uncountable. Existence of equilibria holds under reasonable
  conditions (i.e., each set of actions is a nonempty compact convex
  subset of Euclidean space, and each player utility is continuous and
  quasi-concave in the player's action), all of which are satisfied by
  the MRF-induced infinite game considered here. (See
  \citealp{fudenberg91}, for more information.)
} 
\begin{align*}
\widetilde{M}'_i(Q_i,Q_{\mathcal{N}(i)}) \equiv \sum_{x_i} \sum_{x_{\mathcal{N}(i)}} \left[ Q_i(x_i) \prod_{j \in
  \mathcal{N}(i)} Q_j(x_j) \right] M'_i(x_i, x_{\mathcal{N}(i)}) + H(Q_i)
\end{align*}
and wants to maximize over its mixed-strategy $Q_i$ given the other player mixed-strategies $Q_j$ for all $j \neq i$. 
\begin{property}
\label{pro:mrfinfgame}
The MRF-induced infinite game defined above is an infinite Gibbs potential game with the same graph $G$ and the following potential over the set of individual (product) mixed strategies
\begin{align*}
\Psi(Q) = & \sum_{C \in \mathcal{C}} \sum_{x_C} \left[ \prod_{j \in C}
            Q_j(x_j) \right] \phi_C(x_C) + H(Q)\\
 = & -KL(Q \parallel P) + Z
\end{align*}
where $Z$ is the normalizing constant for $P$.
From this we can derive that the individual player mixed-strategies $\{ Q_i \}$ are a ``pure strategy'' equilibrium of the infinite game if and only if
\[
\textstyle
\KL(Q \parallel P) = \min_{Q'_i} \KL( Q'_i Q^{\times}_{-i} \parallel P).
\]
\end{property}
Or, in other words, {\em if $Q$ is a PSNE of the infinite game, then $Q$ is also a local optimum (or critical point) of the mean-field approximation of $P$.} 

\begin{remark}
\label{rem:lg}
The local payoff function defined above for the infinite game also has
connections to the game theory literature on {\em learning in
  games}~\citep{fudenberg99}. This area studies properties of
processes by which players ``learn'' how to play in (usually repeated)
games; especially properties related to the existence of convergence of
the learning (or playing) dynamics to equilibria. In particular, the
local payoff function is similar to that used by {\em logistic
  fictitious play}, a special version of a ``learning'' process called
{\em smooth fictitious play}. The difference is that the last entropy
term involving the individual player's mixed strategy has a
regularization-type factor $\lambda > 0$ such that players play strict
best-response as $\lambda \rightarrow 0$. In addition, logistic
fictitious play is an instance of a learning process that, if followed
by a player, achieves so called approximate {\em universal
  consistency\/} (i.e., roughly, in the limit of infinite play, the
average of the payoffs obtained by the player will be close to the
best obtained overall during repeated play, {\em regardless of how the
  other players behave}), also known as \emph{Hannan consistency}~\citep{hannan1957}, for appropriate values of $\lambda$ depending on the desired approximation level.

Indeed, it is not hard to see that in fact the best-response
mixed-strategy $Q_i$ of player $i$ to the mixed strategies
$Q_{\mathcal{N}(i)}$ of their neighbors is $Q_i(x_i) \propto$
\begin{multline*}
\textstyle
\exp\left(\sum_{x_{\mathcal{N}(i)}} \left[ \prod_{j \in \mathcal{N}(i)} Q_j(x_j) \right] M'_i(x_i,x_{\mathcal{N}(i)})\right)
\textstyle
=\\ \exp\left(\sum_{C \in \mathcal{C}_i, C \neq \{i\}} \left[ \prod_{j \in C - \{i\}} Q_j(x_j) \right]\phi_C(x_i, x_{C - \{i\}})\right) .
\end{multline*}
Hence, running {\em sequential\/} best-response dynamics in the MRF-induced infinite game is equivalent to finding a variational mean-field approximation via recursive updating of the first derivative conditions.~\footnote{In particular, the process is called a {\em Cournot adjustment with lock-in\/} in the literature on learning in games~\citep{fudenberg99}.} The process will then be equivalent to minimizing the function $F(Q) \equiv \KL(Q \parallel P)$ by axis-parallel updates. The resulting sequence of distributions/mixed-strategies monotonically decreases the value of $F$ and is guaranteed to converge to a local optimum or a critical point of $F$. Hence, the corresponding learning process is guaranteed to converge to a PSNE of the \emph{infinite} game, which is in turn an \emph{approximate} MSNE of the \emph{original} game. But this is not surprising in retrospect, given the last property (Property~\ref{pro:mrfinfgame}). That property essentially states a broader property of {\em all\/} potential games: they are isomorphic to so called {\em games with identical interests}~\citep{Monderer_and_Shapley_1996}, which are games where every player has exactly the same payoff function.

\end{remark}


\begin{remark}
The previous discussion suggests that we could use appropriately-modified versions of algorithms for MSNE, such as {\bf NashProp}~\citep{ortizandkearns03}, as heuristics to obtain a mean-field approximation of the true marginals. 

Going in the opposite direction, the discussion above also suggests that, by treating any (graphical) potential game as an MRF, for any fixed $\lambda > 0$, logistic fictitious play in any potential game converges to an approximate $(\lambda/\min_i |A_i|)$-MSNE of the potential game. Indeed, there has been recent work in this direction, which explores the connection between learning in games and mean-field approximations in machine learning~\citep{rezek08}. That work proposes new algorithms based on fictitious play for simple mean-field approximation applied to statistical (Bayesian) estimation. 
 
The game-induced MRF is a $\lambda$-temperature Gibbs measure. As we take $\lambda \to 0$, we get the limiting $0$-temperature Gibbs measure which is a probability distribution over the set of global maxima of the potential function of the game, and $0$ probability everywhere (i.e., the support of the limiting distribution is the set of joint-actions that maximize the potential function). The support of the $0$-temperature Gibbs measure is a subset of the ``globally optimal'' PSNE of the potential game. But there might be other equilibria corresponding to local optima (or critical points) of the potential function. 

Are there other connections between the Nash equilibria of the game and the support of the limiting distribution? 
\end{remark}

\subsubsection{Correlated Equilibria and \\
Higher-order Variational Approximations} 

\citet{kakade03} designed polynomial-time algorithms based on
linear programming for computing
CE in standard graphical games with tree graphs.
The approach and polynomial-time results extend to
graphical games with bounded-tree-width graphs and 
graphical polymatrix games with tree graphs. \citet{AISTATS07_OrtizSK} (see
also~\citealp{ortizetal06}) proposed the
principle of maximum entropy (MaxEnt) for equilibrium selection of CE
in graphical games. They studied several properties of the MaxEnt CE,
designed a monotonically increasing algorithm to compute it, and
discussed a learning-dynamics view of the algorithm. \citet{ICML2011Kamisetty_594}
employed advances in approximate inference methods 
to propose
approximation algorithms to compute CE. In all of those cases, the
general approach is to use ideas from probabilistic graphical models
to design algorithms to compute CE. The focus of this paper is
the opposite direction: employing ideas from game theory to design
algorithms for belief inference in probabilistic graphical models.

Property~\ref{prop:cevar} suggests that we can use the CE for the MRF-induced game as a heuristic approximation to higher-order variational approximations. In fact, one would argue that in the context of inference, doing so is more desirable because, in principle, it can lead to better approximations that can capture more aspects of the joint distribution than a simple mean-field approximation would alone. For example, mean-field approximations are likely to be poor if the MRF is multi-modal. Motivated by this fact, \citet{jaakkola97} suggest using mixture of product distributions to improve the simple variational mean-field approximation. 

\subsubsection{Some Computational Implications}

But, consider the algorithms of \citet{papadimitriou05_ce} or~\citet{Jiang2015347} (see also \citealp{papadimitriou08}, and~\citealp{DBLP:journals/sigecom/JiangL11}), which we can use to compute a CE of the MRF-induced game in polynomial time. Such CE will be, by construction, also a {\em (polynomially-sized) mixture of product distributions}. (In the case of~\citeauthor{DBLP:journals/sigecom/JiangL11}'s algorithm it will be a mixture of a subset of the joint-action space, which is equivalent to a probability mass function over a \emph{polynomially-sized} subset of the joint-action space; said differently, a mixture of product of indicator functions, each product corresponding to particular outcomes of the joint-action space.) Hence, the algorithms of \citeauthor{papadimitriou05_ce} and~\citeauthor{Jiang2015347} both provide a means to obtain a heuristic estimate of a local optimum (or critical point) of such a mixture {\em in polynomial time.}
The result would not be exactly the same as that obtained by \citet{jaakkola97} in general, because of the extra entropic term mentioned in the discussion earlier. {\em Can we find alternative versions of the payoff matrices, and/or alter Papadimitriou's algorithm, so that the resulting correlated equilibria provides an exact answer to the approximate inference problem that uses mixtures of product distributions?} Regardless, at the very least one could use the resulting CE to initialize the technique of \citet{jaakkola97} without specifying an {\em a priori\/} number of mixtures.

Having said that, both \citeauthor{papadimitriou05_ce}'s and~\citeauthor{Jiang2015347}'s algorithms make a polynomial number of calls to the ellipsoid-algorithm, or more specifically, its ``oracle,'' to obtain each of the product distributions whose mixture will form the output CE. It is known that the ellipsoid algorithm is slow in practice. \citet{papadimitriou05_ce}, \citet{papadimitriou08}, and~\citet{Jiang2015347} leave open the design of more practical algorithms based on interior-point methods. 

Finally, this connection also suggests that we can (in principle) use any learning algorithm that guarantees convergence to the set of CE (as described in the section on preliminaries on game theory where the concept was introduced) as a heuristic for approximate inference. Several so-called ``no-regret'' learning algorithms satisfy those conditions. Indeed, we use two simple variants of such algorithms
in our experiments. Viewed that way, such learning algorithms would be similar in spirit to stochastic simulation algorithms with a kind of ``adaptivity'' reminiscent of the work on adaptive importance sampling (see, e.g.,~\citealp{cheng00,ortiz00_ais,ortiz02_phd},
and the references therein).
Establishing a possible stronger connection between learning in games, CE, and probabilistic inference seems like a promising direction for future research. In fact, as previously mentioned (at the end of Remark~\ref{rem:lg}),
there has already been some recent work in this direction, but specifically for MSNE and mean-field approximations~\citep{rezek08}.

Later in this paper, we present the results of an experimental
evaluation of the performance of a simple no-regret learning algorithm
in computational game theory~\citep{fudenberg99,blum_and_mansour_07,hart00} in the context of probabilistic
inference. Those are iterative algorithms like many other approximate
inference methods such as mean field and other variational
approximations, but closer in spirit to sampling/simulation-based methods such as the Gibbs sampler
and other similar MCMC methods. Indeed, the running time per iteration
of those algorithms is roughly the same as
that of sampling-based methods. We delay the details until the Experiments section (Section~\ref{sec:exp}). 


\subsection{Other Previous and Related Work}

Earlier work on the so called ``relaxation labeling" problem in AI and computer vision \citep{rosenfeld76,miller91} has established connections to polymatrix games \citep{janovskaja68} (see also \citealp{hummel83}, although the connection had yet to be recognized at that time). That work also establishes  connections to inference in Hopfield networks, dynamical systems, and polymatrix games~\citep{miller91,zucker01}. A reduction of MAP to PSNE in what we call here a GMhG was introduced by \citet{yu95} in the same context (see also \citealp{berthod96}); although they concentrate on pairwise potentials, which reduce to polymatrix games in this context. Because, in addition, the ultimate goal in MAP inference is to obtain a {\em global\/} optimum configuration, \citet{yu95} proposed a Metropolis-Hastings-style algorithm in an attempt to avoid local minima. Their algorithm is similar to simulated annealing algorithms used for solving satisfiability problems, and other local methods such as WalkSAT~\citep{selman96} (see, e.g.,~\citealp{russell03} for more information). The algorithm can also be seen as a kind of learning-in-games scheme~\citep{fudenberg99} based on best-response with random exploration (or ``trembling hand'' best response). That is, at every round, some best-response is taken with some probability, otherwise the previous response is replayed. \citet{zucker01} presents a modern account of that work. The connection to potential games, and all its well-known properties (e.g., convergence of best-response dynamics) does not seem to have been recognized within that literature. Also, none of the work makes connections to higher-order (i.e., beyond mean-field) inference approximation techniques or the game-theoretic notion of CE.

\subsection{Approximate Fictitious Play in a Two-player \\Potential Game for Belief Inference in Ising Models}
\label{sec:fp}

This section presents a game-theoretic fictitious-play approach to estimation of node-marginal probabilities in MRFs. The approach this time is more global in terms of how we use the whole joint-distribution for the estimation of individual marginal probabilities. The inspiration for the approach presented here follows from the work of~\citet{1459045}. The section concentrates on Ising models, an important, special MRF instance from statistical physics with its own interesting history. 
\begin{definition}
An \emph{Ising model} wrt an undirected graph $G=(V,E)$ is an MRF wrt $G$ such that 
\[
\P_{\theta}(x) \propto \exp\left(\sum_{i \in V} b_i x_i + \sum_{(i,j) \in E} w_{i,j} x_i x_j\right)
\]
where $\theta \equiv (\b,\W)$ is the set of node biases $b_i$'s and edge-weights $w_{ij}$'s, which are the parameters defining the joint distribution $\P_{\theta}$ over $\{-1,+1\}^n$.
\end{definition}
It is fair to say that interest on more general classes of MRFs originates from the special class of Ising models. It is also fair to say that, because of the relative simplicity and importance of Ising models for problems in statistical physics, as well as to other ML and AI applications areas such as computer vision and NLP, Ising models have become the most common platforms in which to empirically study approximation algorithms for arbitrary MRFs. In short, simplicity of presentation and empirical evaluation guide the focus of Ising models in this section: Generalizations to arbitrary MRFs are straightforward but cumbersome to present. 
Hence, in this manuscript, we omit the details of such generalizations.

As an outline, the current section begins with an algorithmic instantiation of the iterative approach. The exact instantiation depends on whether we are using CE or MSNE as the solution concept. The section then follows with 
an informal discussion of the game-theoretic foundations of the general framework behind the approach, and a discussion of immediate implications to computational properties and potential convergence.

Denote by $\T_G$ the set of all spanning trees of connected
(undirected) graph $G=(V,E)$ that are maximal with respect to $E$
(i.e., does not contain any spanning forests). If spanning tree $T \in
\T_G$, we denote by $E(T) \subset E$ the set of edges of $T$. To
simplify the presentation of the algorithm, let
\[
\Mtilde_{\Tcal}(\mu,T) \equiv \sum_{(i,j) \in E} \indicator{(i,j)
  \in E(T)} \, w_{ij} \mu_{(i,j)}
\] 
and 
\[
\Psi_{X,\Tcal}(x,T) \equiv \sum_{i \in V} b_i x_i +
\sum_{(i,j) \in E} \indicator{(i,j) \in E(T)} w_{ij} x_i x_j . 
\]

Initialize $x^{(1)} \gets \text{Uniform}(\{-1,+1\}^n)$, and for each $(i,j) \in E$, $\widehat{\mu}^{(1)}_{(i,j)} \gets x^{(1)}_i x^{(1)}_j$.
At each iteration $l = 1,2,\ldots,m,$
\begin{algorithmic}[1]
\State \label{alg:mst_step} $\T^{(l)} \gets \argmax_{T \in \T_G} \Mtilde_{\Tcal}(\widehat{\mu}^{(l)}_{(i,j)} ,T)$
\State $T^{(l)} \gets \text{Uniform}\left(\argmax_{T \in \T_G} \T^{(l)}\right)$
\State $s_l \gets \text{Uniform}(\{1,\ldots,l\})$
\State $\mathcal{X}^{(l+1)} \gets \argmax_{x \in
    \{-1,+1\}^n} \Psi_{X,\Tcal}(x,T^{(s_l)})$
\State $x^{(l+1)}  \gets \text{Uniform}\left(\mathcal{X}^{(l+1)}\right)$
\ForAll{$(i,j) \in E$}
\State $v^{(l+1)}_{(i,j)} \gets x^{(l+1)}_i x^{(l+1)}_j \times \begin{cases}
 1, & \text{if MSNE,}\\
\indicator{(i,j) \in E(T^{(s_l)})}, & \text{if CE} \\
\end{cases}$
\State $\widehat{\mu}^{(l+1)}_{(i,j)} \gets \frac{l \; \widehat{\mu}^{(l)}_{(i,j)} + v^{(l+1)}_{(i,j)}}{l+1}$
\EndFor
\end{algorithmic}
For each Ising-model's random-variable index $i = 1,\ldots,n$, set \[ p^{(m+1)}_i = \frac{1}{m+1} \sum_{l=1}^{m+1} \indicator{x^{(l)}_i = 1}\] as the estimate of the exact Ising-model's marginal probability $p_i \equiv \mathbf{P}(X_i=1)$.

The running time of the algorithm is dominated by the computation of
the maximum spanning tree (Step~\ref{alg:mst_step}) which is $O(|E| +
n \log{n})$. All other steps take $O(|E|)$.

Within the literature on probabilisitic graphical models,
\citet{Hamze:2004:FT:1036843.1036873} propose an MCMC approach based
on sampling non-overlapping trees. While our approach has a sampling
flavor, its exact connection to MCMC is unclear at best. Also, the
spanning trees that our algorithm generates may overlap.

The following discussion connects the algorithm above to an approximate version of fictitious play from the literature on learning in games in game theory. 
For the most part, we omit discussions to approximate variational
inference in this manuscript, except to say that TRW
message-passing~\citep{1459045} is the inspiration behind our proposed
algorithm above.

The game implicit in the heuristic algorithm above is a two-player
potential game between a \emph{``joint-assignment'' (JA)} player and a
\emph{``spanning-tree'' (ST)} player. The potential function is
\(\textstyle \Psi_{X,\Tcal}(x,T) . \)
The payoff functions $M_X$ and $M_\Tcal$ of the JA player and the ST player, respectively, are identical and equal the potential function $\Psi_{X,\Tcal}(x,T)$: formally, \(\textstyle M_X(x,T) = M_\Tcal(x,T) = \Psi_{X,\Tcal}(x,T) \). Note that the payoff function of the ST player is \emph{strategically equivalent} to the function \(\textstyle \sum_{(i,j) \in E} \indicator{(i,j) \in E(T)} w_{ij} x_i x_j . \)

Technically, this is a game with identical payoffs, which are known to have what~\citet{Monderer1996258} called the \emph{fictitious play property}: the empirical play of fictitious play is guaranteed to converge to an MSNE of the game. While determining a best-response for the ST player is easy (i.e., using an algorithm for computing maximal spanning tree such as Kruskal's, as we do in our implementation for the experiments), unfortunately the same is in general not possible for the JA player, whose best-response is as hard as computing a MAP assignment of another Ising model with the same graph and (generally non-zero) bias\/node parameters, but a slightly different set of edge-weights.~\footnote{As mentioned earlier, there are some instances for which this computation is actually possible in polynomial time. In fact, this would have been possible for the type of Ising models with planar two-dimensional grid graph, also known as a ``square lattices,'' we used in the experiments, if we would have chosen those models to have zero biases, or the edge-weights had some special characteristics. Unfortunately, there is no guarantee that the specific Ising models randomly drawn would satisfy those conditions in general. As we discuss shortly, we settle for a simple computation of the best-response for the JA player using stochastic fictitious play~\citep{fudenberg99}.} 

One approach to deal with the problem of obtaining a best-response
from the JA player is to draw one tree uniformly at random from the
empirical distribution and find a best-response to that tree. Such an
approach is equivalent to a type of smooth best-response. If both
players were to do the same, \emph{simultaneously}, the result is a
stochastic version of fictitious play or \emph{stochastic fictitious
  play} for short~\citep{fudenberg99}. The empirical distribution of
play of stochastic fictitious play in a game with identical payoffs,
or what's strategically equivalent, any potential game, also converges
to an MSNE of the game~\citep{10.2307/3081987}. In our case, however,
we really have a type of ``hybrid'' sequential-version, where the ST player is always behaving as in standard fictitious play, while the JA player is behaving according to a stochastic fictitious play. 

In addition, as an alternative to the best-response computation for
player JA, we might want to add an entropic (preference) function of
the mixed-strategy to the JA player as an additional term in JA's
payoff, so that the result is really a ``smooth'' best-response, or more specifically in this case a \emph{smooth stochastic fictitious play}~\citep{fudenberg99}. Such an addition would make the connection to variational inference more evident, and would allow us to develop more direct bounds on the quality of the variational result. The main problem is that we do not know of any study of such hybrids within game theory. In addition, most instances of fictitious play assume \emph{simultaneous} moves. Numerical instability is another problem we found in practice when using such smooth best-response. Even in instances where that was not a problem, the performance was indistinguishable, in a formal statistical sense, from the version of the algorithm that we propose above.

In the context of belief inference, we believe it actually makes more sense to have a so called ``sequential'' play, where players trade moves: the JA player starts by choosing some action (i.e., full, joint assignments to the random variables), the ST player best-responds to that action, then the JA player best-responds to the ST player's action, continuing in that way, such that at each round, each player is best-responding to the \emph{empirical} distribution of play~\footnote{In game theory, this is also known as the \emph{belief} distribution of play each player has about the others' future mixed-strategy based on previously observed play.}  up to the time the player makes a move (i.e., draws an action). While this type of sequential process often helps to stabilize the dynamics and improve the likelihood of convergence, it seems that such sequential processes have received considerably less attention than their simultaneous-move counterpart within the game-theory community. 

We conjecture, however, that the type of fictitious play process defined above in fact converges. We believe that the proof follows from combining results from standard and stochastic fictitious play for games with identical payoffs, which are (strategically equivalent) instances of potential games~\citep{Monderer1996258,10.2307/3081987}. The derivation is complex and not trivial, involving key mathematical concepts from the literature in stochastic approximation. Delving into such level of complexity not only goes beyond the scope of this paper, but more importantly, doing so distracts attention from the paper's main focus: to provide a general, broad illustration of how ideas and results from game theory may be useful in providing alternative, effective, and practical approaches to hard belief-inference problems in probabilistic graphical models. Thus, we leave the formal proof of our conjecture as future work.

As a last point, it is important to understand and keep in mind that, as it is well-known, in the context of potential games, while \emph{sequential best-reply} converges to a PSNE (i.e., a joint assignment), \emph{fictitious play} can converge to an MSNE of the game.~\footnote{Recall that in fictitious play, each player uses the \emph{empirical distribution of play} as an estimate or belief of how the other player would behave in the future, \emph{not just} the other player's \emph{last action} as in sequential best-reply.} \citet{Monderer1996258} provide an example in a 2-player 2-action normal-form (coordination) game with identical payoffs. Said differently, the resulting empirical distribution of play for the JA player may be to what~\citet{Monderer1996258} themselves call a ``purely mixed strategy'' (i.e, every action is played with positive probability; or said differently, the corresponding probability mass function has full suport over the action set of the player).~\footnote{Other names used in game theory are \emph{totally mixed strategy} or \emph{mixed strategy with full support.}} In the context of belief inference, the resulting mixed-strategy would correspond to an (approximate) marginal distribution, not a particular joint-assignment. Hence, in the context of belief inference, the convergence of the procedure above may not have to be to a single (possibly local) optimum of the potential function $\Psi_{X,\Tcal}$: in principle, convergence could be to a (non-deterministic) \emph{mixture} over joint-assignments. In fact, this is what we observe in our experiments, albeit after only a finite number of iterations. A thorough understanding of the convergence properties observed in practice requires considerably more experimental work than is reasonable within the context and purpose of the work described in this manuscript.

\subsection{Sketch of algorithm derivation and relation to TRW}
The connection between TRW and the algorithm presented in this section
results from a stochastic minimization of a precise upper bound on a
variational approximation that uses a joint distribution (CE) or a
product distributions (MSNE) over the spanning trees of the MRF graph
$G=(V,E)$ and the original MRF's random variables $X$.
Here is a sketch of the precise mathematical expressions for the
CE-based case. (The ones for the MSNE-case are very similar, and
omitted for brevity.) While the derivation is more general, we only
present it in the context of Ising models.

We use the following notation for the purpose of the discussion
here. Let $Q_{X,\Tcal}$ be the
  variational joint distribution over the random variables $X$ and
  $\Tcal$ corresponding to joint variable assignments and spanning
  tree, respectively. Let $Q_X$ be the marginal probability of
  $Q_{X,\Tcal}$ over $X$: i.e., $Q_X(x) \equiv \sum_{T \in
  \T_G} Q_{X,\Tcal}(x,T)$, for all $x \in \{-1,+1\}$. Let $P_X \equiv P_\theta$ be the
ground-truth joint distribution (defining the Ising model) we would
like to approximate. Denote by \[ \KL(Q_X || P_X) \equiv \sum_x Q_X(x)
\ln\frac{Q_X(x)}{P_X(x)} \] the KL-divergence of between $Q_X$ and
$P_X$ with respect to $Q_X$. Let
$\widehat{Q}_{X,\Tcal}$ the \emph{empirical} joint distribution of "joint
actions" for the both players generated during fictitious play: i.e.,
$\widehat{Q}_{X,\Tcal}(x,T) \equiv \frac{1}{m} \sum_{l=1}^{m}
\indicator{x^{(l)} = x, T^{(l)} = T}$. Let $v_{ij} \equiv
\E_{\widehat{Q}_{X,\Tcal}}\left[X_i X_j\right] = \sum_x \widehat{Q}_X(x) x_i x_j$, where
$\widehat{Q}_X(x) \equiv \sum_{T \in \T_G} \widehat{Q}_{X,\Tcal}(x,T)$ denotes the empirical marginal over $X$
only; that is, summed over all spanning trees $T$ over $G=(V,E)$ with
respect to $\widehat{Q}$, which is clearly easy to compute:
$\widehat{Q}_X(x) \equiv \sum_{T \in \T_G} \widehat{Q}_{X,\Tcal}(x,T)$
for all $x$. Let 
\begin{align*} 
u_{ij} \equiv & \E_{Q^*_{X,\Tcal}}\left[ \indicator{(i,j) \not\in \Tcal} X_i
                X_j\right] \\
= & \textstyle \sum_{x,T} Q^*_{X,\Tcal}(x,T) \indicator{(i,j) \not\in T} x_i x_j \; .
\end{align*}
where $Q^*_{X,\Tcal} \in
\argmax_{Q_{X,\Tcal}} \sum_{x,T} Q_{X,\Tcal}(x,T) \Psi(x,T)$.
Denote by $H(Q^*_X) \equiv - \sum_x Q^*_X(x) \ln Q^*_X(x)$ the standard
"Shannon's entropy" of $X$ with respect to the marginal of $Q^*_{X,\Tcal}$ over
$X$ (i.e., $Q^*_X(x) = \sum_T Q^*_{X,\Tcal}(x,T)$); and similarly for $H(\widehat{Q}_X)$.
After some algebra, we can obtain the following bound for the variational approximation:
\begin{align*}
\min_{Q_X} \KL(Q_X || P_X) \leq & - \max_{Q_{X,\Tcal}} \sum_{x,T} Q_{X,\Tcal}(x,T) \Psi(x,T) + \\
& - \sum_{(i,j) \in E}
       w_{ij} u_{ij}  - H(Q^*_X) + \ln Z \; .
\end{align*}
The first term in the bound, maximizing over $Q_{X,\Tcal}$, inspires the application of fictitious play.
As an aside, note that we can generate a family of upper bounds (details omitted); e,g,, for $\widehat{Q}_X$, 
\begin{align*}
\min_{Q_X} \KL(Q_X || P_X)
\leq & - \sum_{(i,j) \in E} w_{ij} v_{ij} - H(\widehat{Q}_X) + \ln Z \; .
\end{align*} 
Note that $H(\widehat{Q}_X)$ is easy to compute, and that the last expression leads immediately to an easily computable lower bound on $\ln Z$.

\section{Experiments}
\label{sec:exp}

In this section we present the results of synthetic experiments on the
performance of the game-theoretic-inspired heuristics we propose in
this paper for approximate belief inference in MRFs. Our algorithms
have very simple implementations. We also compare them with the most
popular approximation algorithms and heuristics, with equally simple implementations, proposed in the literature on probabilistic graphical models. 

\subsection{Experimental Design: Synthetic Models}

The experimental design in terms of the class of Ising models is as
in~\citet{NIPS2013_4964}. We consider Ising models with $d \times d$
simple grid graphs, which are planar (i.e., no ``wrap around'' edges,
such that each of the four corner nodes have exactly two neighboring
nodes, any other non-internal node has exactly three neighbors, while
the rest, i.e., all internal nodes, have exactly four
neighbors). Hence, the number of variables or nodes is $n=d^2$. 
We used $d \in \{8,12\}$ for our experiments.
We did not consider edge-weights magnitude parameters $1.0$ or $1.5$,
because it is really hard to beat a Gibbs sampler for maximum weight
magnitudes smaller than $2.0$, relative to the bias parameters $b_i$'s
being in the real-valued interval $[-1,1]$. The reason for this might
be that, because, as stated in~\citet{NIPS2013_4964}, the mixing rate
of a Gibbs sampler in such models grows roughly exponential with the
magnitude, the induced Markov chain mixes pretty fast for such cases;
thus convergence is quick. For each value $w \in \{2.0,2.5,3.0,3.5,4.0\}$, we generated random Ising
models with edge-weights $w_{ij} \sim \text{Uniform}([-w,w])$ or
$w_{ij} \sim \text{Uniform}([0,w])$ for the ``mixed'' or
``attractive'' case, respectively, for each $(i,j)\in E$, i.i.d., and
node biases $b_i\sim \text{Uniform}([-1,+1])$, also i.i.d. for all $i$, and
independent of the edge-weights.

One exception on the class of Ising models used for evaluation is a
class we use with edge-weights with constant \emph{magnitude} (i.e., $w = \max_{(i,j) \in E} |w_{ij}|$), but in
which we vary the probability $q$ of attractive edge-weights; that is,
given a probability $q$, the sign of the edge-weight are i.i.d. random
variables in which the sign is positive with probability $q$, and
negative with probability $1-q$.~\footnote{The weight of each edge
  $(i,j) \in E$ is a random variable of the form $W_{ij} = (2 S_{ij}
  -1) w$, where the $S_{ij} \sim \text{Bernoulli}(q)$, i.i.d., and $w$
  is a positive constant.}
We propose this class of Ising
models for future evaluations of approximate belief inference
techniques. For evaluation using this class, we consider $w\in
\{2.0,2.5,3.0,4.0\}$. For each $q$, we randomly
generated $50$ Ising models as samples for $w=4$, and $5$ samples for
each $w\neq 4$.

Note that despite the graphs being planar, the bias parameter is non-zero in general, so that the known polynomial-time exact algorithms for planar graphs do not technically apply. 

Here, we consider simple \emph{no-regret algorithms} from the
literature of learning in games~\citep{fudenberg99, blum_and_mansour_07}. The two most common 
notions of regret are \emph{external} and \emph{swap}
regret, the latter being stronger than the former. (Another name often used for external regret is
``unconditional regret,'' while other names for swap regret are
``internal regret'' and ``conditional regret.'' We refer the reader to
the references to standard literature on learning in games for specific
definitions.) There are several
such no-regret algorithms in the literature with different types of convergence
guarantees depending on the exact notion of regret used.  Here we consider two types of no-regret algorithms, and leave the evaluation of
other no-regret algorithms for future work. One type of algorithm we
consider is really a class of algorithms based on the 
\emph{Multiplicative Weight Update (MWU)}
algorithm~\citep{blum_and_mansour_07}. In our implementation of the
MWU algorithm, for each player $i$ at each round $t \geq 1$, we set the probability of
playing action $x_i$ at round $t+1$, which we denote by $x_i^{(t+1)}$, to be $p_{t+1}(x_i) \propto p_t(x_i)
\left( 1 - \eta_t (1 - \bar{M}_i(x_i,x_{-i}^{(t)}))\right) $, where
$\eta_t$ is analogous to a learning rate in ML (i.e., the step size when using
standard gradient descent/ascend for optimization), and
$\bar{M}_i$ is the normalized payoff function for player $i$ (i.e.,
the expression $x_i \left( \sum_{(i,j) \in E} w_{i,j} x_j + b_i \right)$ normalized so that minimum and
maximum values are $0$ and $1$, respectively). In general, if we set the value of
$\eta_t$ to a constant $\eta$ we guarantee that the empirical joint
distribution over joint actions induced by the played sequence of
joint actions $x^{(t)}$ converges to the set of
approximate CCE, where the level of the approximation depend on
$\eta$. 
If we set $\eta_t = \sqrt{\frac{\ln(2)}{t}}$ then we guarantee
that convergence to the set of CCE.~\footnote{The set of CCE is a superset of CE
and is related to the concept of external regret. Indeed, convergence
to the set of CCE implies that each player has no external
regret, and we say that the empirical play of the player is ``Hannan
consistent''~\citep{hannan1957}, or equivalently, ``universally
consistent''~\citep{fudenberg99}.} There is a simple construction that allows us to
use the MWU algorithm to construct algorithms for which a player can 
have either no swap regret, or approximately swap-regret, depending on 
the value of $\eta_t$: if constant ($\eta$) the empirical
distribution of empirical play converges to the set of approximate CE,
while if set as above convergence is guaranteed to
the set of CE.~\footnote{The set of CE is related to the concept of
  swap regret. Indeed, convergence
to the set of CE implies that each player has no swap
regret.} We refer to the different versions of no-regret
algorithms based on MWU related to convergence to exact or approximate CCE as
``mw\_er'' and ``mw\_er\_cf,'' respectively; and to those related to exact or
approximate CE as ``mw\_sr'' and ``mw\_sr\_cf,'' respectively.
We set $\eta$ to $0.01$ in all of our experiments. 

We also evaluate a
simple (approximate) no-swap-regret algorithm by~\citet{hart00}, which we denote
as ``nr'' from now on. Like all (approximate) no-swap-regret algorithms, nr guarantees to converge to (the
\emph{set} of approximate) CE. Each iteration
of nr takes roughly the same amount of time as that for Gibbs
sampling. We set the number of iterations of the nr algorithm to
$10^5$ for the standard experimental setup, and to $10^6$ for our
proposed new evaluation setting. Our exact implementation is a natural
adaptation we believe is more amenable to the belief-inference
setting. In particular, we evaluate a version in which we update the
mixed-strategy each player uses to draw an action at every iteration
$t$ as follows. For each player, (1) we set the probability of
switching the player's last action being equal
to the empirical regret, or $0$ if the empirical regret is
negative; and (2) we set the player's probability of playing action $+1$
by ``damping''  the currently
suggested probability of playing $+1$, $p_t(1)$, for the corresponding
player by the original algorithm:~\footnote{The algorithm determines its
  suggested probability solely on the positively-truncated empirical regret.}
that is, we use the update $0.99 
\times p_t(1) + 0.01 \times (0.5)$.
We also use $10^5$ iterations. Also, we only present results for the sequential, ``semi-stochastic'' fictitious play we 
discuss in Section~\ref{sec:fp}, for the case of CE only, which we denote as ``fp (ce)'' from now on. We set the number of iterations $m=15$.~\footnote{That number of iterations is relatively low, but given that our implementation is in MATLAB, setting $m=15$ is roughly the number of iterations for which the amount is roughly the same as that for our C implementation of TRW, as described in~\citet{1459045}, but without optimizing for the parameters $\rho_{ij}$'s, which we set to a constant $=0.55$ for all edges $(i,j) \in E$. Clearly this is an unfair comparison for fp (ce). The optimization of $\rho_{ij}$'s involves performing a maximum spanning tree computation at each iteration until convergence, and this operation follows each TRW message-passing with fixed $\rho_{ij}$'s. While such an optimization is tractable, 
and optimizing for the $\rho_{ij}$'s does seem to improve the
upper-bounds on the log-partition function, it is not clear from the
experimental results in~\citet{1459045} that the improvement on the
quality of the individual marginal estimates justify the extra work
necessary for the optimization.} Finally, the results for the
MSNE-instantiation of the fictitious play algorithm we propose are
quite similar to those for fp (ce), at least for $m=15$; thus, we omit
those results in the interest of keeping the plots less ``crowded''
and thus easier to interpret. 

We compare the different mw-type algorithms, the simple nr algorithm, and our proposed fp (ce) to (1) standard
mean-field approximation (mf), with sequential\/axis-parallel updates;
(2) standard belief propagation (bp), with simultaneous updates; (3)
TRW (trw); and (4) the Gibbs sampler (gs). The running times per
iteration of all
methods is $O(|E|)$, except that of fp (ce) which is $O(|E|+n
\log{n})$, and of course that of bl which is constant. In the next paragraph, we provide more detail
on the specifics of the implementations of methods (1--4).

As baseline (bl), we use the simplest possible estimator from the
perspective of average marginal-error to measure quality: always use
0.5 as the estimate of the exact marginal distribution of each
variable. Certainly, one would expect that for an algorithm to be
competitive, its performance should be better than bl. As we soon
discuss, our experimental results suggest that this is not always the
case; that is, several standard methods, including some of the ones
proposed here and even state-of-the-art such as TRW, do not satisfy
that condition for ``hard'' cases. We evaluate mean field (mf) using
sequential\/axis-parallel updates, stopping if the maximum absolute
difference in probability values between iterations is $\leq 10^{-5}$,
and using a maximum number of iterations $=10^6$. For belief propagation (bp) we use simultaneous updates, and ``smooth'' the update based on the average of the current value and the new value in order to ``dampen'' or at least try to prevent oscillations and improve the likelihood of convergence,~\footnote{It is well known that bp may not converge in MRFs with loopy graph, such as the Ising model with grid graph we are using here for our experiments.} stopping if the maximum absolute difference in probability values between iterations is $\leq 10^{-7}$, and a maximum number of iterations $=10^5$. For tree reweighed message-passing (trw), we use a constant parameter $\rho=0.55$ for all corresponding edge-appearance-probability parameters $\rho_{ij}$'s~\citep{1459045}, along with a smooth update and the same stopping criterion as for bp. For the Gibbs sampler, we use $10^6$ iterations.

\subsection{Experimental Results: Synthetic Models}

\begin{figure*}[h]
\vspace{1in}
\begin{center}
\includegraphics[width=0.5\linewidth]{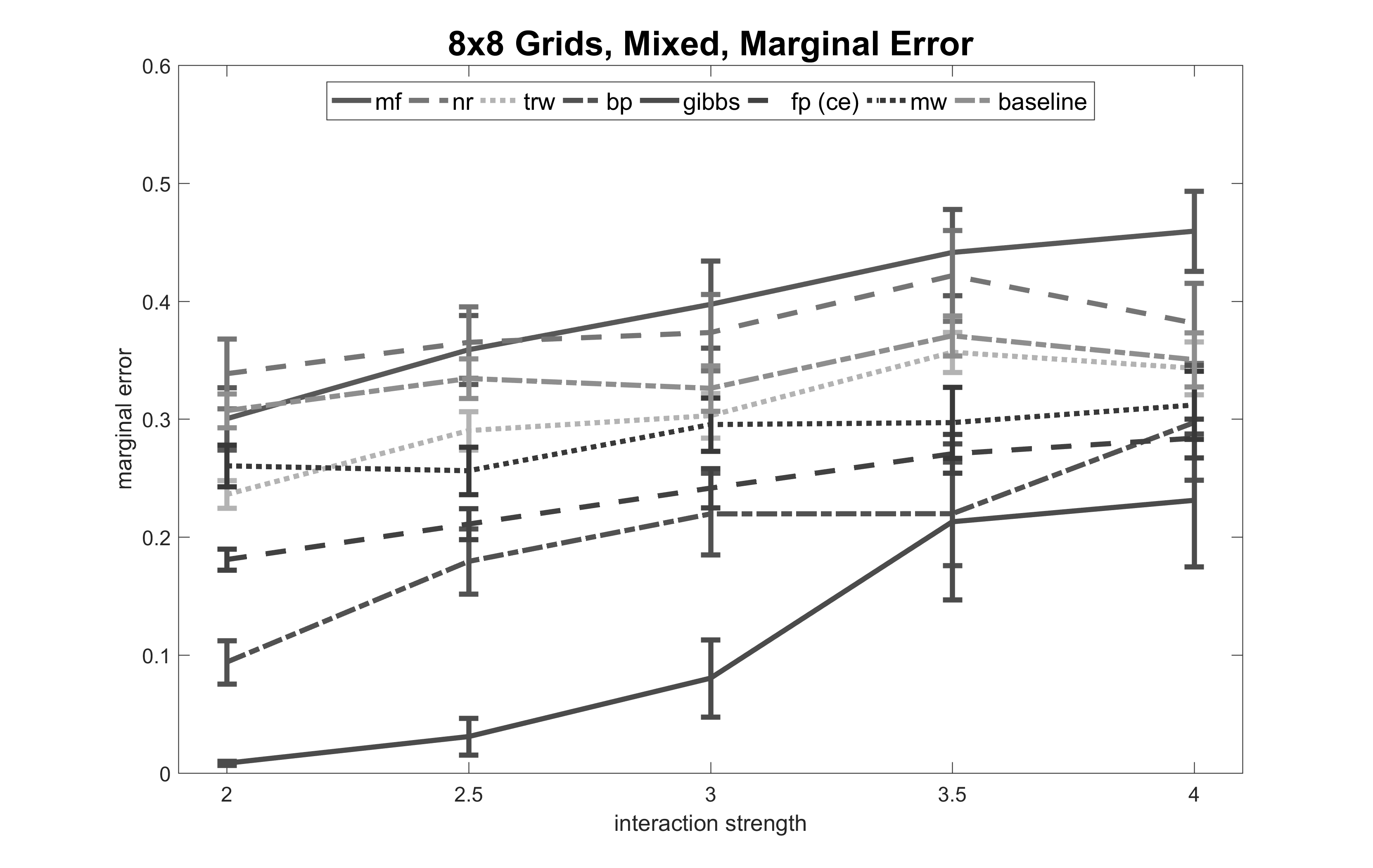}\includegraphics[width=0.5\linewidth]{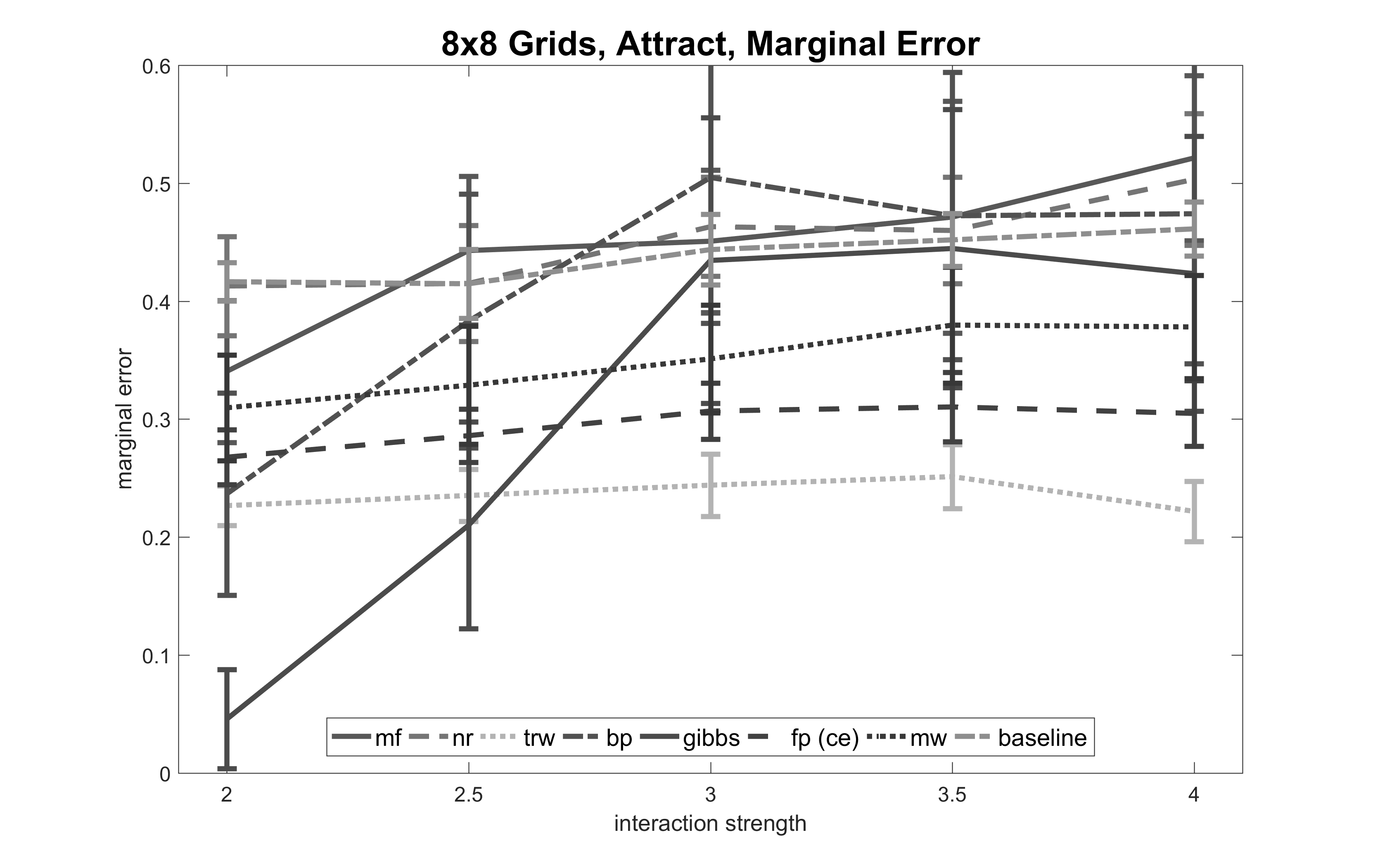}
\includegraphics[width=0.5\linewidth]{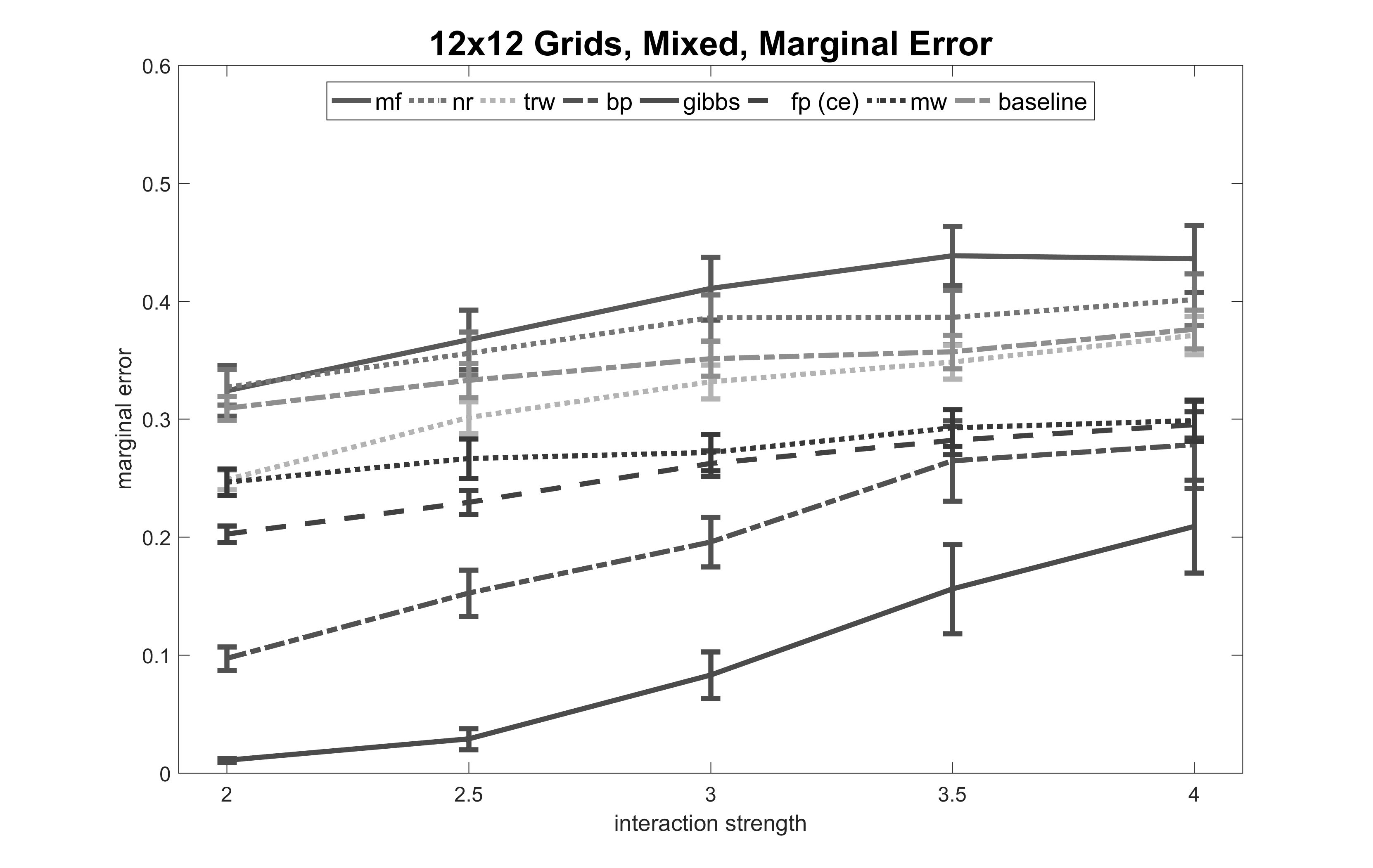}\includegraphics[width=0.5\linewidth]{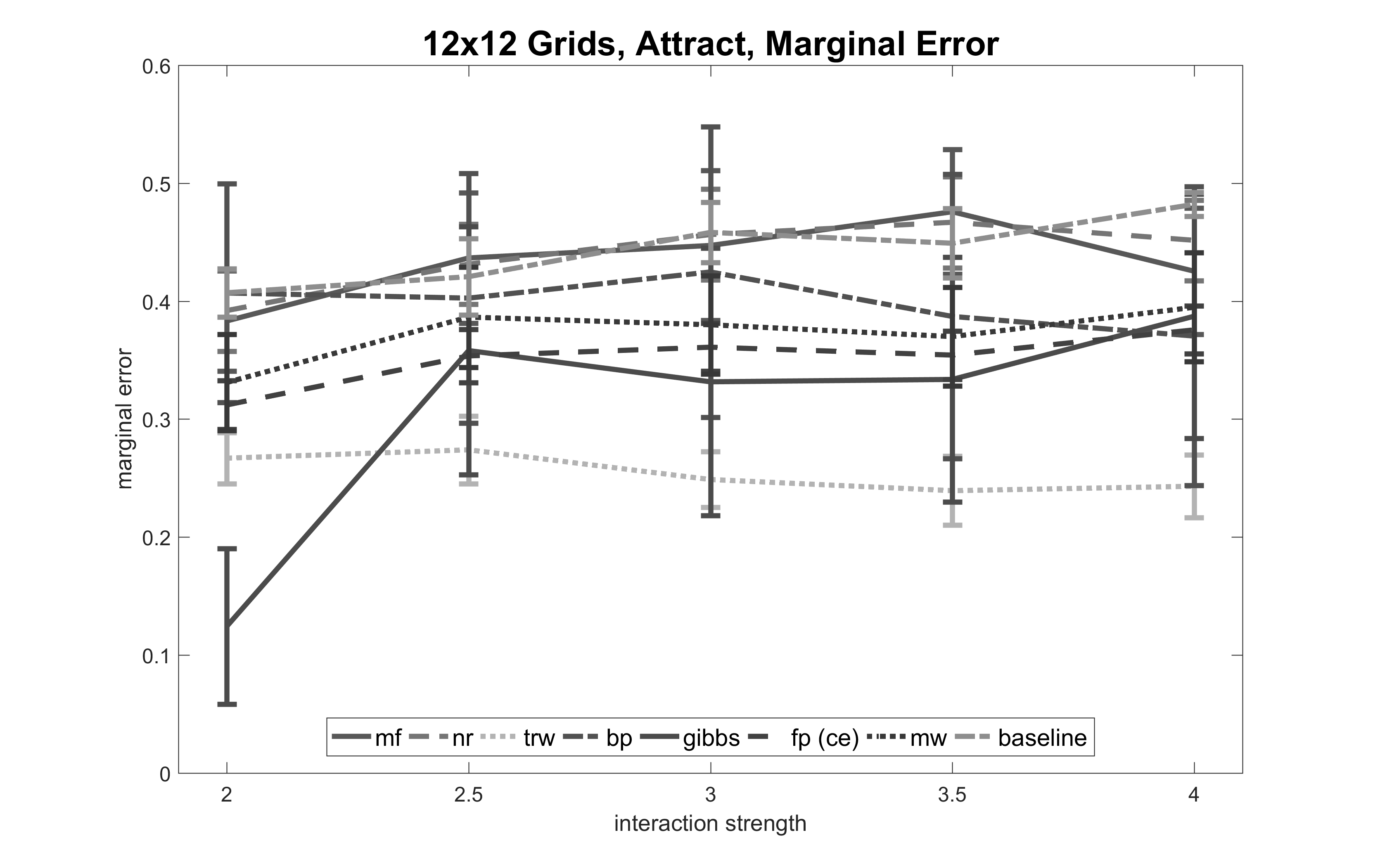}
\end{center}
\caption{{\bf Standard Evaluation on Ising Models with 8x8 and 12x12 Grids.} The
  left and right plots are for the so-called ``mixed'' and
  ``attractive'' instances of Ising models, respectively. For all
  plots, the x-axis is the largest magnitude of the edge-weights: i.e., $w =
  \max_{(i,j) \in E} |w_{ij}|$. The y-axis is the average, over $50$
  randomly generated Ising models, of the \emph{average}, over all of
  the $144$ variables, of the absolute difference between the estimate
  and exact marginal probability for the random variable corresponding
  to that node, along with their corresponding 95\% confidence
  intervals (CIs). 
The legend in each plot is for different approximation algorithms: bl = baseline; mf = mean field; bp = belief propagation; trw = tree reweighed message-passing; nr = simple no-regret algorithm; gs = Gibbs sampler; mw = multiplicative weights; and fp (ce)  = the CE version of our version of the fictitious play for the 2-player potential game described in Section~\ref{sec:fp}. We refer the reader to the main body for implementation details and a thorough discussion of the results.}
\label{fig:std_eval}
\end{figure*}

Fig.~\ref{fig:std_eval} summarizes our results for the most common
classes of Ising models considered in the experimental evaluation of
approximation algorithms and heuristic for belief inference in the
literature as described above. 
We perform hypothesis testing for the result in these classes of Ising
models using paired z-tests on the individual (i.e., not joint)
differences, each with p-value $0.05$. Hence, all the statements are
statistically significant with respect to such hypothesis tests.
 Note that there is no globally best approximation technique overall
 for these classes. Finally, plots for both 8x8 and 12x12 models are included
to illustrate how the relative performance of these approximation algorithms are
not strongly affected by the grid size $d$. For brevity, we will only discuss  
results for the 12x12 case. We refer the reader to Appendix~\ref{app:8x8} for experimental results and discussion for the 8x8 case.

Among all mw-type algorithms, we only
present the results of mw\_er\_cf because it outperforms all other types almost
consistently, as we discuss later in this section and plot in
Fig.~\ref{fig:mw_compare}. Hence, we refer to mw\_er\_cf simply as 'mw' from now on.

\paragraph{``Mixed'' 12x12 case (Bottom left plot, Fig.~\ref{fig:std_eval}).} 
Clearly, gs is 
 best for all $w$ in this case. Among the other approximation
 algorithms, we observe the following.
\begin{enumerate}
\item fp (ce) is worse than bp for $w<3.5$, and indistinguishable from bp for $w\geq3.5$.
\item fp (ce) is consistently better than trw.
\item trw is consistently worse than bp.
\item mw is worse than fp (ce) for $w<3.0$, and indistinguishable from fp (ce) for $w\geq3.0$
\item All methods, except for mf and nr, are consistently better than
  bl. mf and nr are consistently worse than bl.
\item mf is better than nr for $w\geq3.0$, and indistinguishable from each other for $w<3.0$.
\end{enumerate}

\paragraph{``Attractive'' 12x12  case (Bottom right plot, Fig.~\ref{fig:std_eval}).}
In this case, there is no clear overall best. We also observe the
following.
\begin{enumerate}
\item trw is best among all methods for $w\geq3.0$, indistinguishable from gs for $w=2.5$, and worse than gs for $w=2.0$.
\item fp (ce) is worse than gs for $w=2.0$, but better than gs for $w=4.0$, and indistinguishable from gs otherwise.
\item mw and fp (ce) are consistently indistinguishable.
\item mf, nr, and bp are consistently indistinguishable from each other, except for $w=2.0$, where bp is better than nr.
\item bp and bl are consistently indistinguishable, except for $w=4.0$, where bp is better.
\end{enumerate}

\begin{figure*}[h]
\vspace{1in}
\begin{center}
\includegraphics[width=0.5\linewidth]{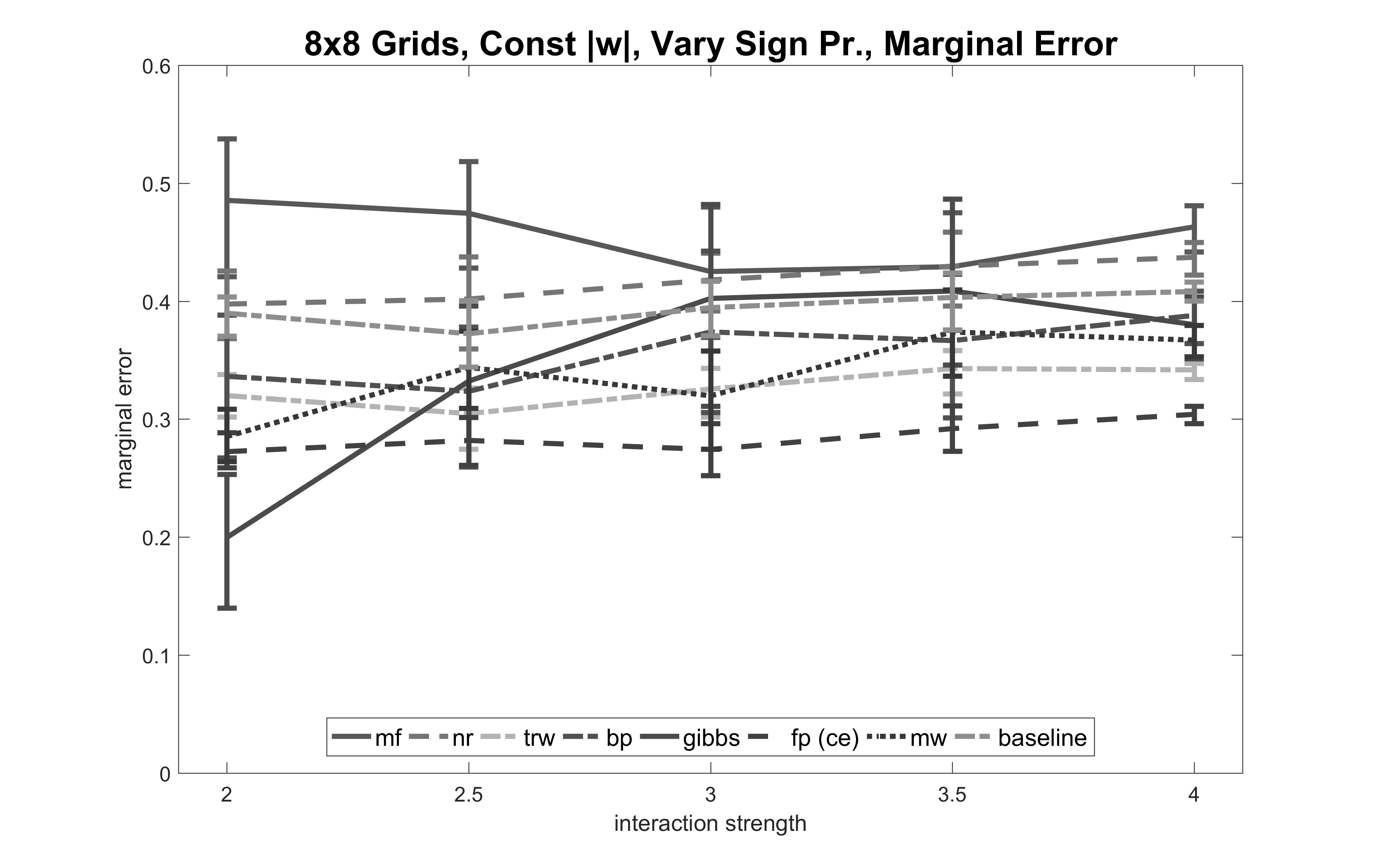}\includegraphics[width=0.5\linewidth]{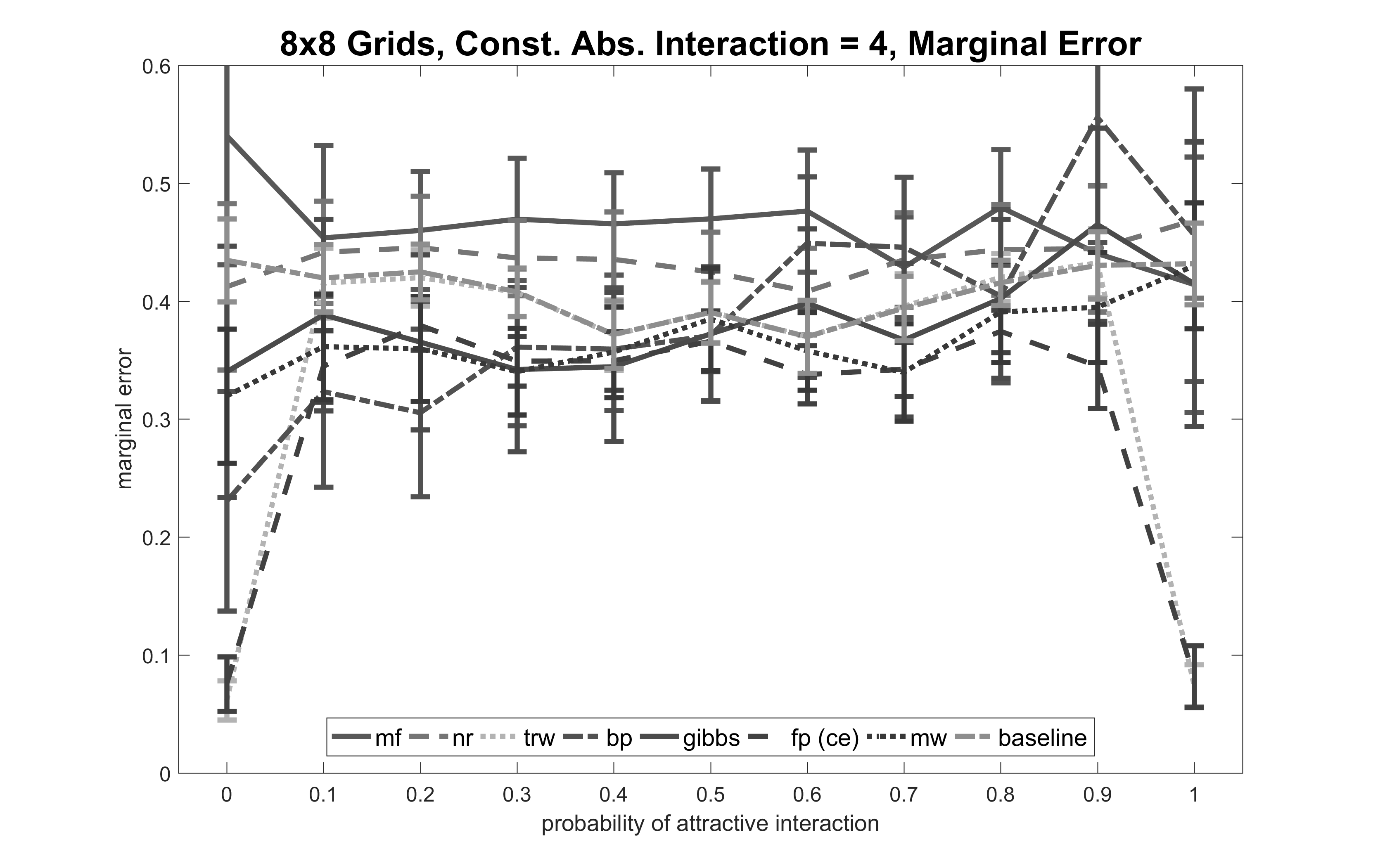}
\includegraphics[width=0.5\linewidth]{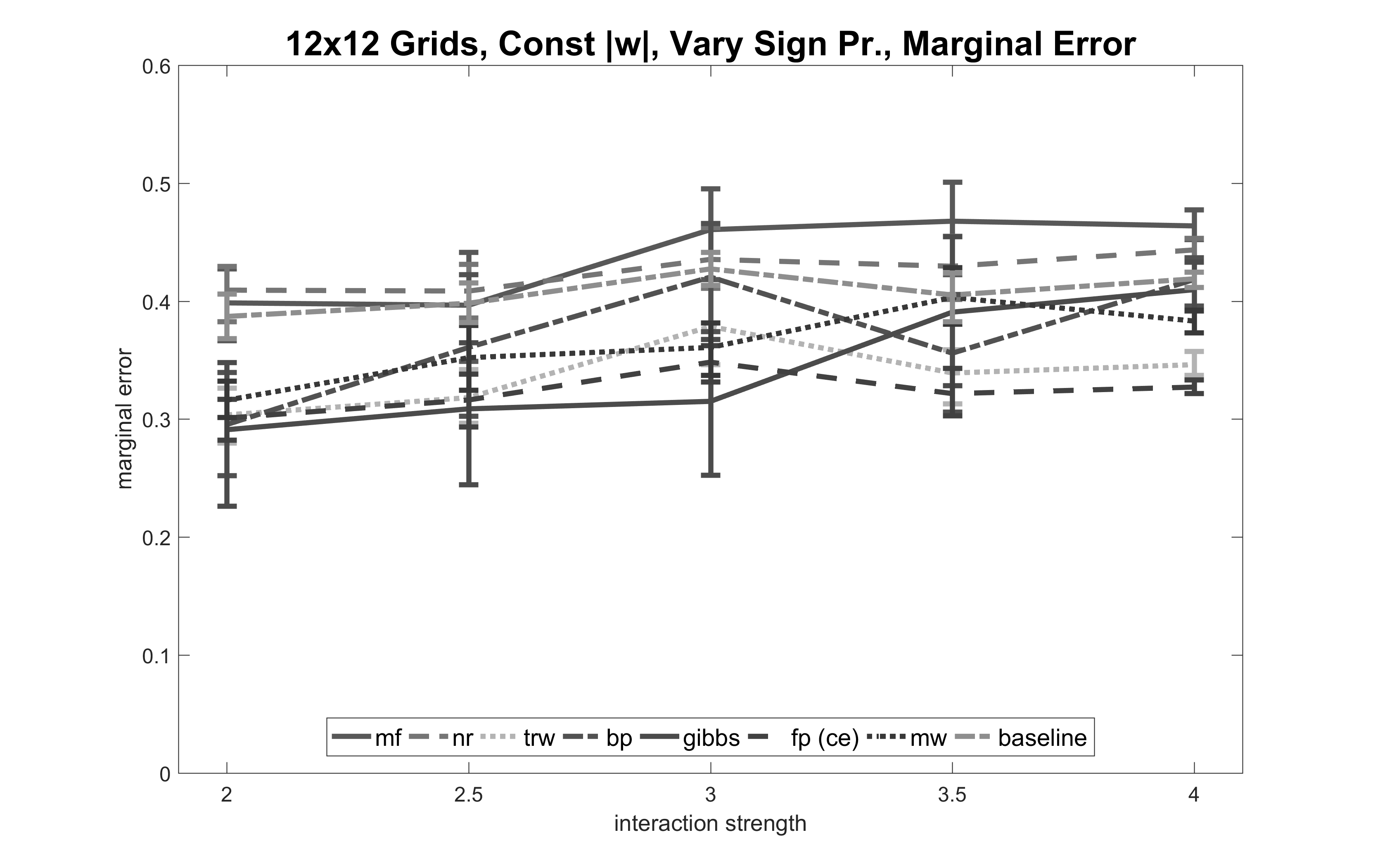}\includegraphics[width=0.5\linewidth]{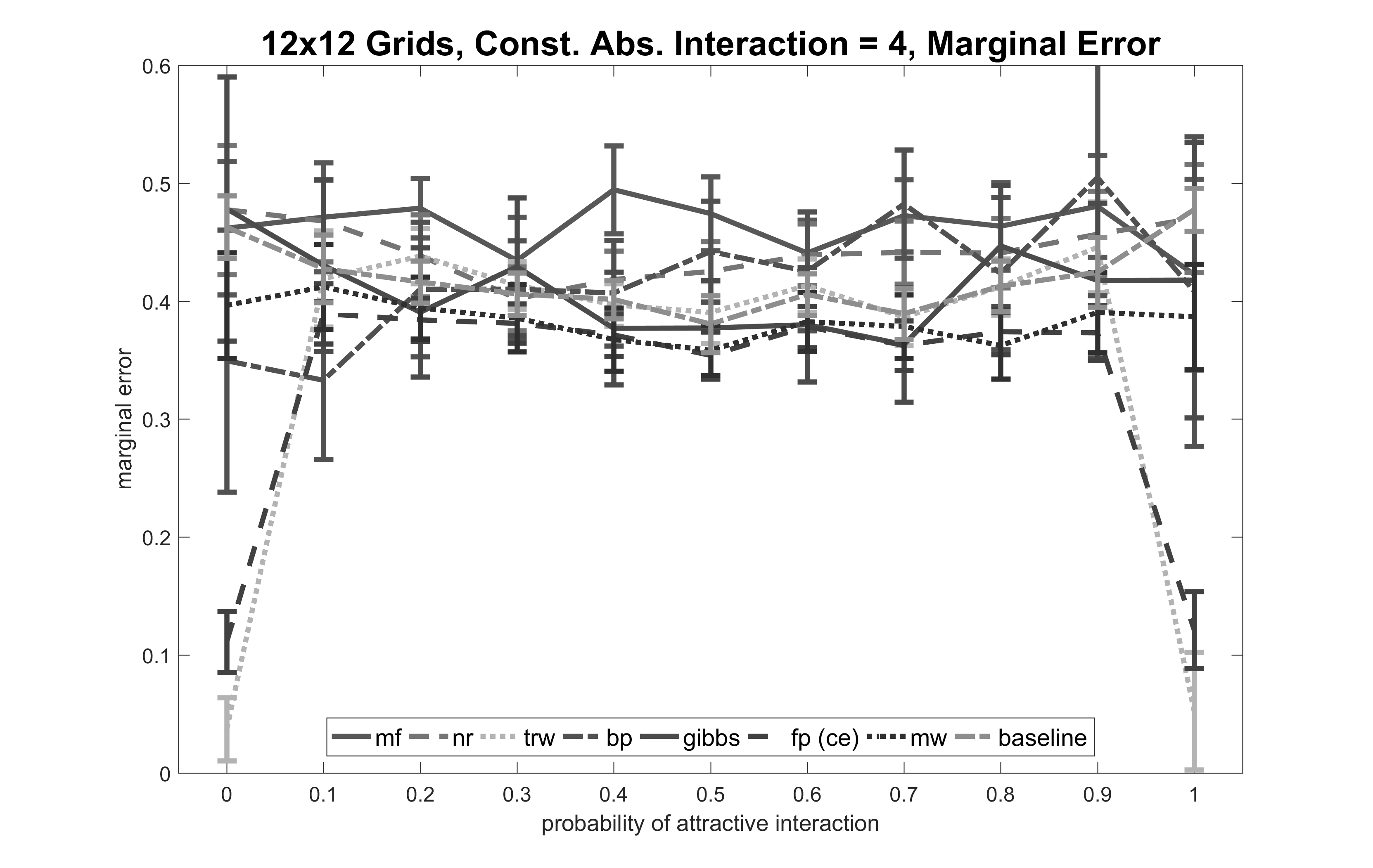}
\end{center}
\caption{\small {\bf Evaluation on Ising Models with 8x8 and 12x12 Grids, Uniform Interaction
  Magnitude, and Varied Probability of Attractive Interactions.}
(\emph{Left plots}) The x-axis, y-axis, and legend are as in
Fig.~\ref{fig:std_eval}, except the edge-weight \emph{magnitude} $w$
is constant for each interaction strength in the x-axis (i.e., $2$,
$2.5$, $3$, and $4$), and nr uses $10^6$ iterations. For all cases,
the result is the average over all values of the probability of
attractive interaction $q \in \{0.0,0.1,\ldots,0.9,1.0\}$, and over
$5$ Ising models for each $q$; \emph{except} for the case of constant
edge-weight magnitude $w=4$, in which case the average for each $q$ is
over $50$ Ising-model samples. Said differently, the overal average
for the cases of $w\in\{2.0,2.5,3.0\}$ is over a total of $55$ Ising
models, while those for the case of $w=4$ is over a total of $550$
models. Note that the standard 95\%
CIs based on a Gaussian approximation resulting from
the Central Limit Theorem (CLT) do not directly apply here because the averages are
over different $q$ values, each of which may have different
distributional properties (e.g., different variances). For $w < 4$,
because we are computing the average marginal-error over every $q$,
each based on only $5$ samples, we use the bootstrap method to compute
the 95\% CIs over the overall average for each method
and each $w$, using $100$ samples. For $w=4$, because we have $50$
samples for each $q$, we use a properly adapted version of the
standard 95\% CIs which modifies the calculation of the overall
variance to account for distributional differences from each $q$. 
(\emph{Right plots}) Results for each $q$ value with $w=4$, with $50$ models as
samples for each, along with their corresponding individual 95\% CIs computed as usual. 
We refer the reader to the main body for a thorough discussion.}
\label{fig:const_w}
\end{figure*}

\begin{figure*}[h]
\vspace{1in}
\begin{center}
\includegraphics[width=0.5\linewidth]{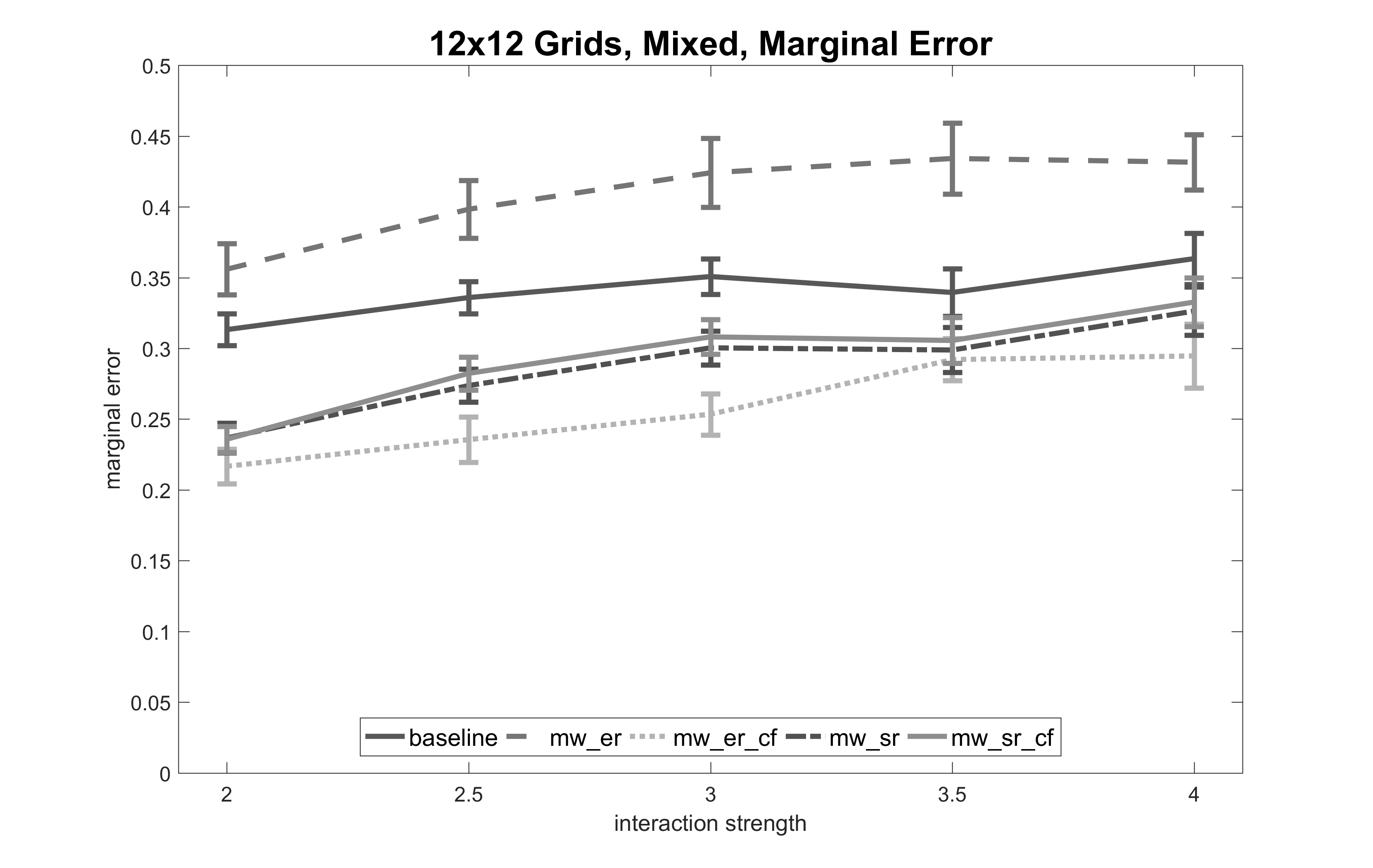}\includegraphics[width=0.5\linewidth]{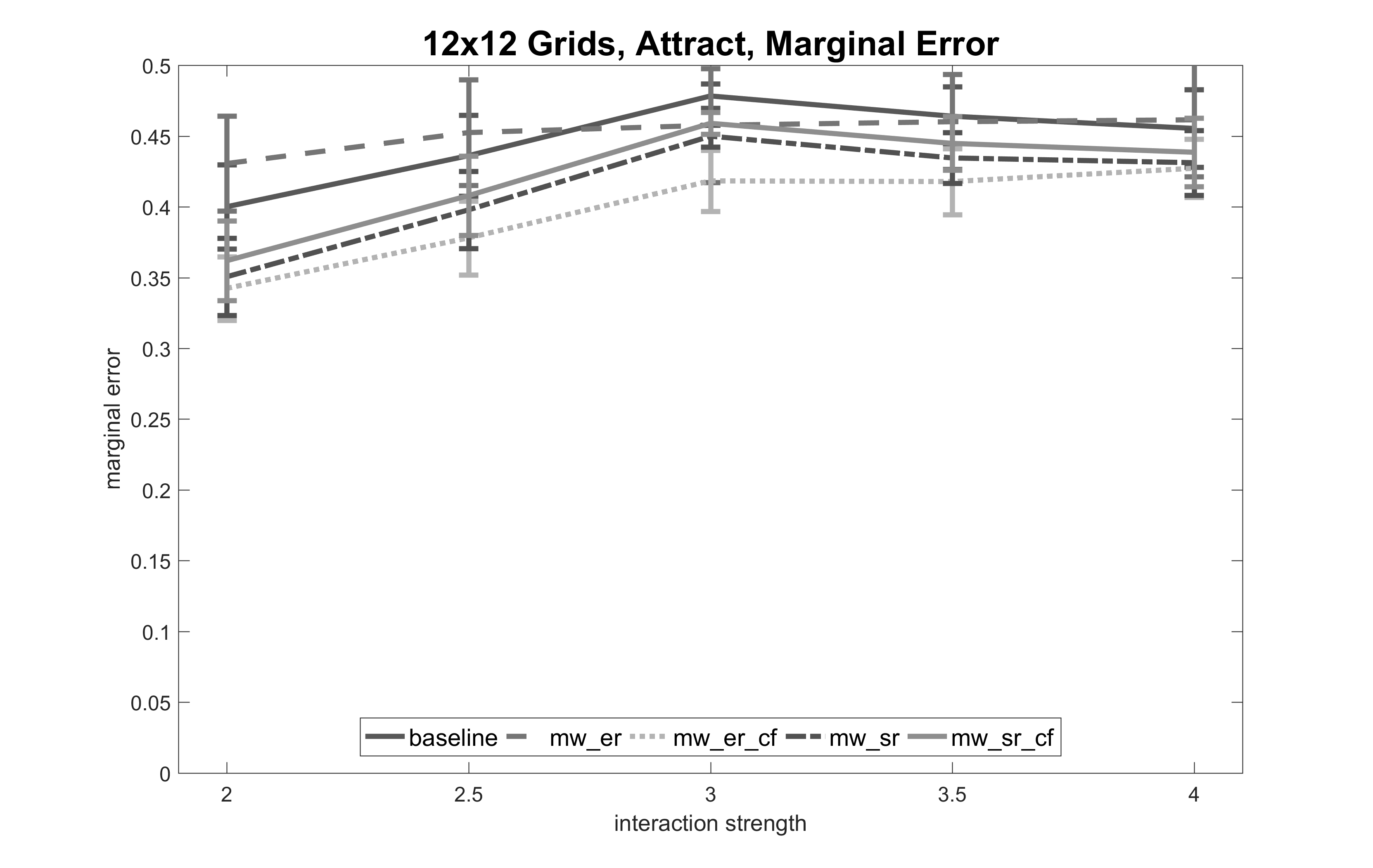}
\includegraphics[width=0.5\linewidth]{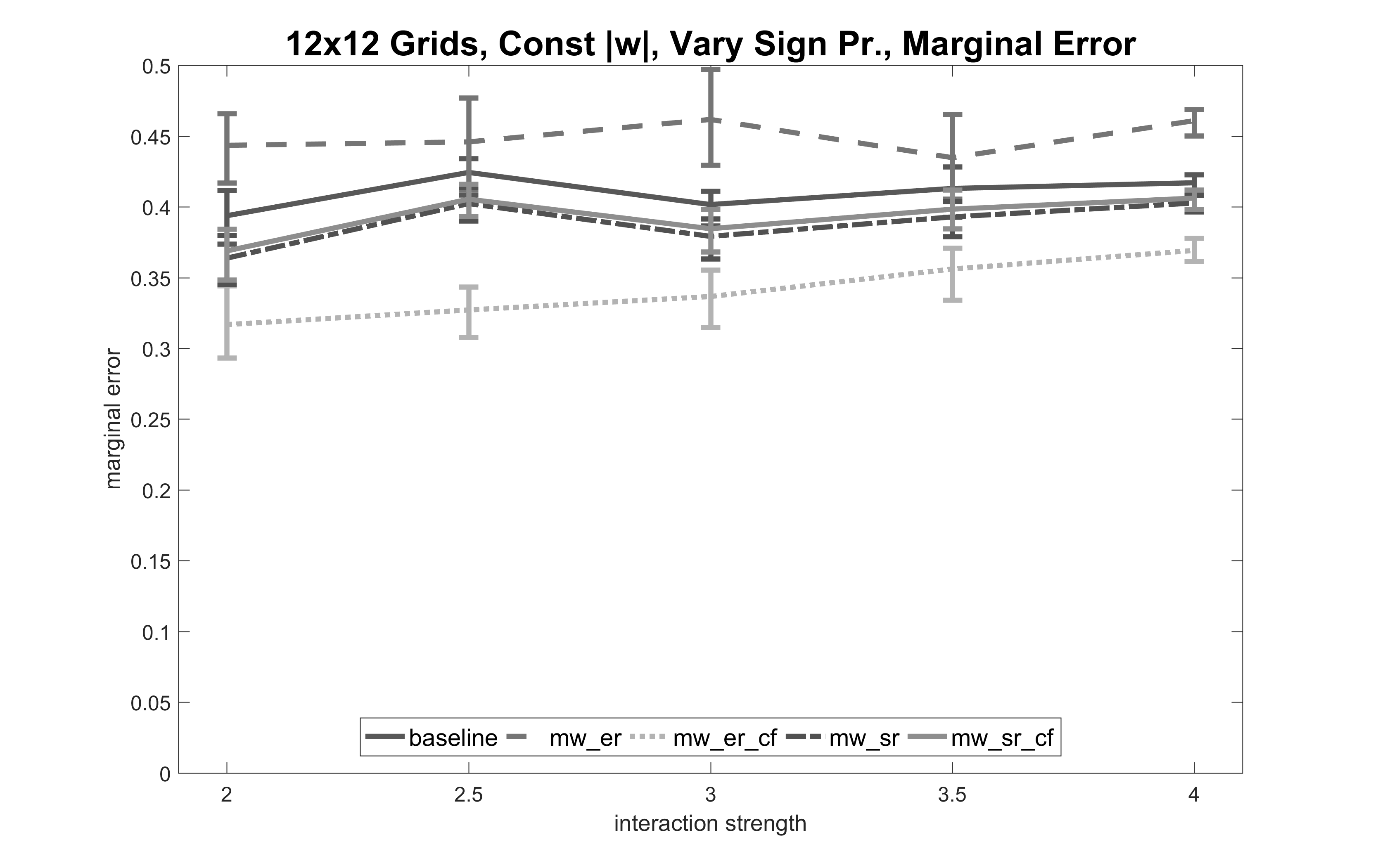}\includegraphics[width=0.5\linewidth]{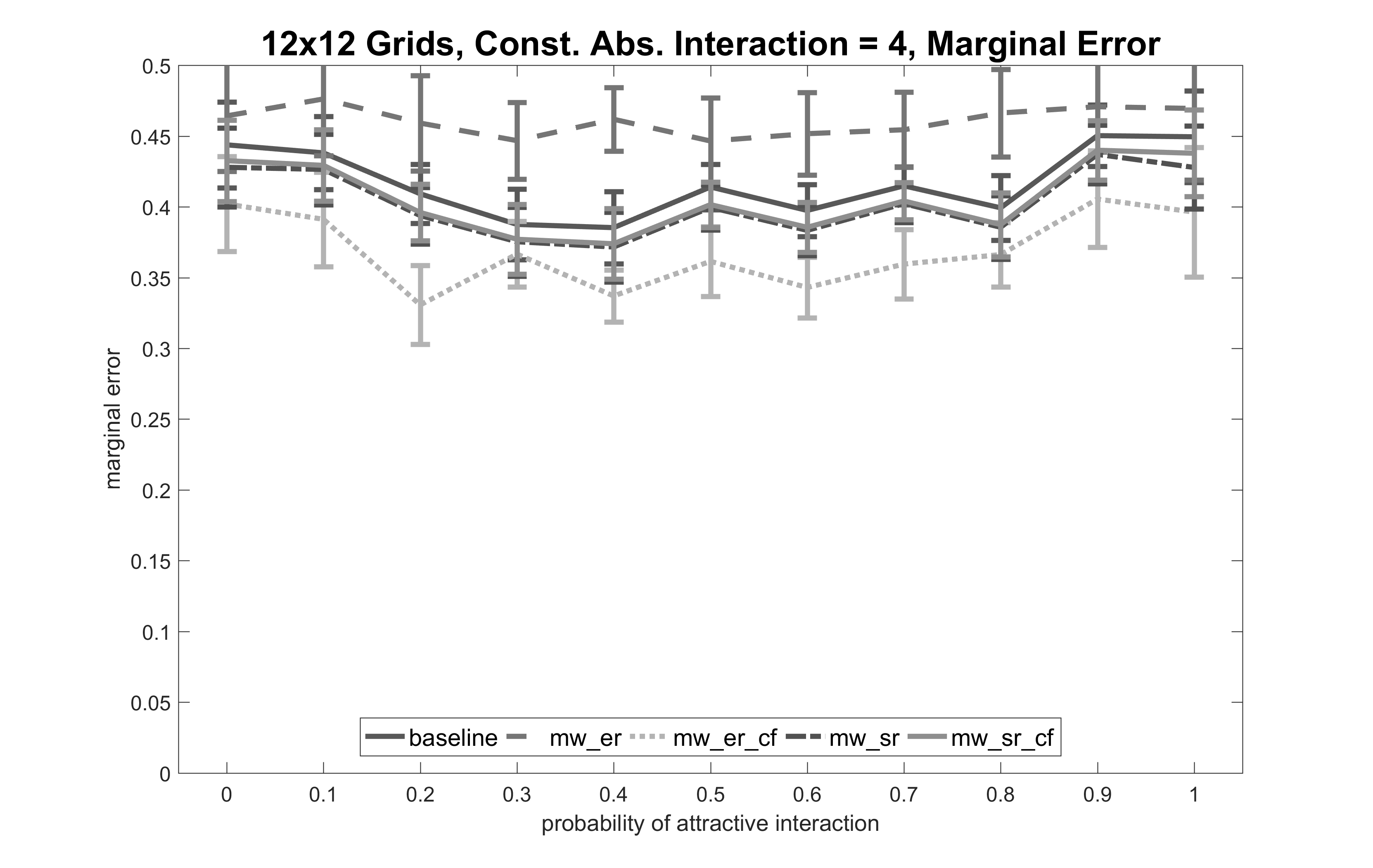}
\end{center}
\caption{{\bf Evaluation on Ising Models with 12x12 Grids, Multiplicative Weights Algorithms.}
The top plots are the same as those in  Fig.~\ref{fig:std_eval} and the bottom plots are the same as those in Fig.~\ref{fig:const_w},
with the exception that only  multiplicative weights algorithms are included in these plots (along with the baseline algorithm).
The legend is as follows: mw\_er = external regret minimization; mw\_er\_cf = external regret minimization using a constant $\eta = 0.01$;
mw\_sr = swap regret minimization; mw\_sr\_cf = swap regret minimization using a constant $\eta = 0.01$. 
We refer the reader to the main body for more details regarding the implementation of these algorithms.
}
\label{fig:mw_compare}
\end{figure*}

Fig.~\ref{fig:const_w} summarizes our experimental results for a class
of Ising models which appears to lead to ``harder'' Ising-model
instances.~\footnote{Such class of models follows from our general
  experience with similar models. We find that instantiating Ising
  model parameters using densities over edge-weights tended to yield
  to relatively easier models than the ones we obtain by fixing the
  magnitude of the edge-weights and varying the probability of their
  sign, independently for each edge.} 
We perform hypothesis testing for the result in these classes of Ising
models using two approaches depending on $w$. For $w=4$, where we draw
$50$ models as samples for each $q$, we use appropriately modified paired z-tests on the individual (i.e., not joint)
differences, each with p-value $0.05$. We modify the calculation of
the variances resulting from the average over the samples computed for
each $q$. We do so because the distributional properties of the empirical
mean\/average for each $q$ may differ. 
For $w < 4$, where we only draw $5$ models as samples for each $q$, we
use bootstrapped-based, individual, paired hypothesis-testing over
each pair of aggregate differences between the methods for each of
those values of $w$; we use $100$ bootstrap samples, and p-value $0.05$
All the statements are
statistically significant with respect to such hypothesis tests.

\paragraph{Aggregate results for 12x12 (Bottom left plot, Fig.~\ref{fig:const_w}).}
The bottom left-hand plot in Fig.~\ref{fig:const_w} shows the aggregate results for
this case. There is no clear
overall best over all $q$. We also observe the
following.
\begin{enumerate}
\item fp (ce) is best among all methods except for when $w=2.0$, where gs is better.
\item trw is second best among all methods, except for when $w=2.0$, where it is third best (behind fp (ce) and gs).
\item bp is consistently better than mf and nr except when $w=3.5$, where it is indistinguishable from nr (but still better than mf).
\item mf is consistently worse than bl, except when $w=4.0$, where they are indistinguishable. 
nr is also consistently worse than bl, except when $w=2.5$, where they are indistinguishable
\item gs is consistently better than mf, nr, and bl, except when $w=4.0$, where gs and bl are indistinguishable.
\item mw is better than bl when $w<3.5$, but indistinguishable from bl when $w\geq3.5$.
\end{enumerate}

\paragraph{Results for constant edge-weight magnitude $w=4$ as a
  function of probability of attractive interaction $q$ (Right plots, Fig.~\ref{fig:const_w}).}
The right-hand plots in Fig.~\ref{fig:const_w} shows finer-grain results
for this case, for both the 8x8 and 12x12 models.
The results suggest that in fact
such instances of Ising models tend to be harder in the sense that even
state-of-the-art algorithms such as trw are no better than the simple
baseline estimation, in which $\widehat{p}_i=0.5$ for all
nodes/variables $i$, for about half of the full range of values of
the sign probability $q$ (i.e., for $q \in
\{0.1,0.4,0.5,0.6,0.8,0.9\}$). In fact, the performance of trw is
\emph{almost exactly the same} as baseline across the range of
non-extreme values of $q$. (Note how the plot of the values for trw
and bl are essentially on top of each other for values of $q$ other than $0$ or $1$.) 
On the other hand, note how fp (ce) is consistently better than bl
across the whole range of values for $q$. In fact, fp (ce) is always in the set of (statistically)
best performers for all $q$ with the exception of $q \in \{0.0, 1.0\}$, where trw is better. 
Almost all the methods other than fp (ce) are no better, and often worse, than bl.
Two notable exceptions are that trw and mw beat bl only when $q \in
\{0.0, 1.0\}$. Also, mw and fp (ce) are indistinguishable, except for $q \in \{0.0, 1.0\}$ where fp (ce) is better.

\begin{figure}[h]
\vspace{1in}
\begin{center}
\includegraphics[width=\linewidth]{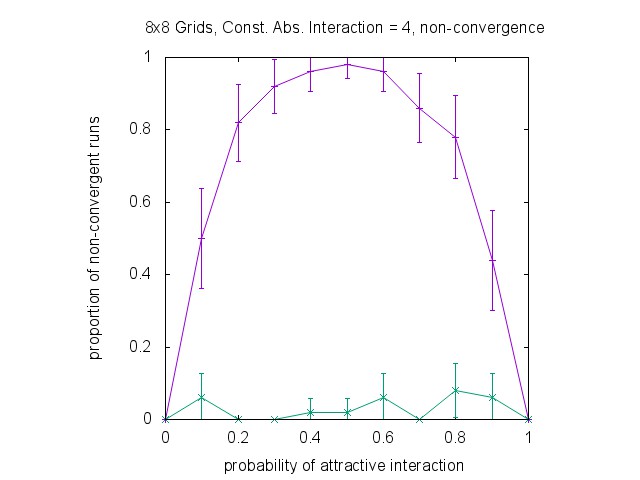}
\end{center}
\caption{{\bf Evaluation on Ising Models with 8x8 Grids, Uniform Interaction
  Magnitude (w= 4), and Varied Probability of Attractive Interactions $q$:
  Proportion of Non-convergent BP and TRW Runs.} This plot shows
proportion, along with standard individual 95\% CIs,
of non-convergent runs (y-axis) of bp (higher curve) and trw (lower
curve), as a function of the probability of attractive interaction $q$
(x-axis) for Ising models with constant edge-weights magnitude equal
to $4$. The setup is as described in the right plot of Fig.~\ref{fig:const_w} for the case of edge-weight magnitude $w=4$. The proportion is out of $50$ runs for each $q \in \{0, 0.1,
0.2,\ldots, 0.9,1\}$. Note how the convergence of bp degrades 
when $q$ nears $0.5$. Note the almost symmetric effect on
non-convergence for bp. Note also that bp non-convergence seems
uncorrelated with its performance, as shown in Fig.~\ref{fig:const_w}
(Right plot).
While trw may also show non-convergence outside non-uniform edge-weights, the effect is less drastic than for bp.}
\label{fig:bp_non_conv}
\end{figure}

Fig.~\ref{fig:bp_non_conv} plots the proportion of non-convergent runs
of bp (higher curve) and trw (lower curve). Note the interesting
behavior of bp: the likelihood of convergence diminishes considerably
as $q$ nears $0.5$.
The effect is almost symmetrical. In contrast, the effect on the
non-convergence of trw is negligible. Note, however, that the bp's
non-convergence does not seem to really affect its performance in
terms of marginal error.
This plot provides additional evidence for our claim that the generative model of random Ising models used for evaluation does lead to harder problem instances.

\paragraph{Results for the effect of different types of no-regret
  algorithms.}
Fig.~\ref{fig:mw_compare} shows the results of various types of no-regret
algorithms based on the multiplicative-weights algorithm. The variants
result from the combination of
(a) external vs. swap regret and
(b) exact vs. approximate no-regret guarantees. The take-home message is that the
version for external regret with approximate no-regret
guarantees, which we refer to both as 'mw' and 'mw\_er' throughout
this section, is consistently better or no worse than the others, as
we stated at the beginning of this subsection (second paragraph). This
result appears counter-intuitive given that such variant has the
lowest guarantees in terms of optimality/equilibrium conditions. Said
differently, the possible set of solutions is largest among all
variants. An analogy with gradient-descent based optimizations in
other machine-learning contexts may provide a
possible explanation for this behavior. For example, it is well known
that reducing the learning rate or step size in inverse proportion
to the number of iterations when learning neural networks via backprop
theoretically guarantees convergence in parameter (weight) space. Yet, it is equally
well-known that doing so is often slow in practice, and that using a
small contant as the learning rate tends to lead to faster convergene
to good models, despite the lack of theoretical guarantees. We refer
the reader to the caption in Fig.~\ref{fig:mw_compare} for further discussion.

\begin{figure}[h]
\vspace{1in}
\begin{center}
\includegraphics[width=\linewidth]{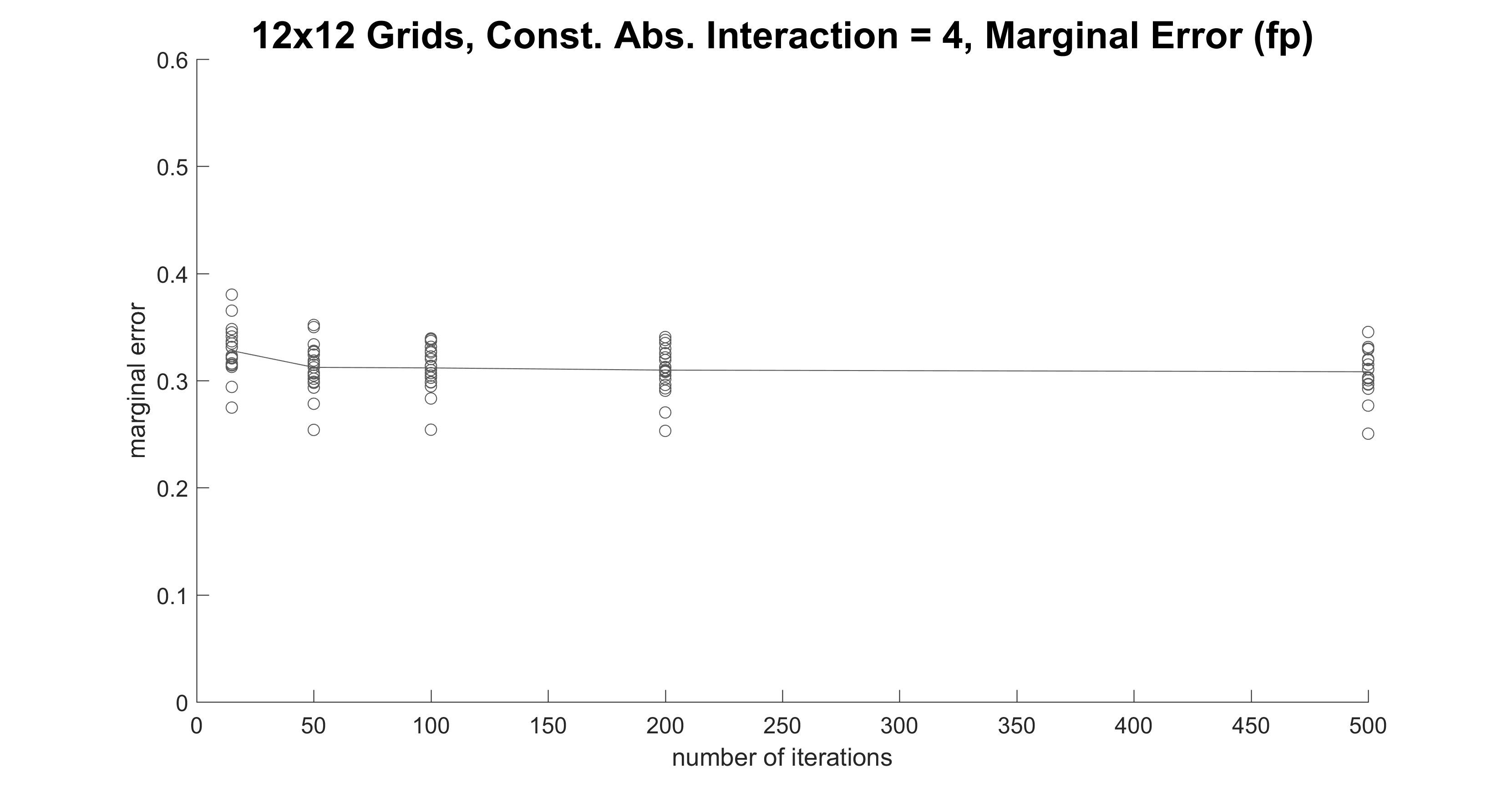}
\end{center}
\caption{{\bf Evaluation on Ising Models with 12x12 Grids, Uniform Interaction
  Magnitude (w= 4), and Varied Probability of Attractive Interactions:
  Marginal Error of fp (ce) by Number of Iterations.}
This plot shows the marginal error of the estimates obtained by the fp algorithm, for
different numbers of iterations used in the algorithm. The x-axis is the number of iterations
$m$, while the y-axis is the same as in Fig.~\ref{fig:std_eval} and Fig.~\ref{fig:const_w}.
The marginal error of each run is represented by a circle on the graph.
Each run is the result of averaging over all values of the probability of attractive interaction 
$q \in \{0.0,0.1,\ldots,0.9,1.0\}$.
The average marginal error is shown as a line, and is obtained from 20 randomly generated Ising 
models and corresponding estimates.
The number of iterations used were $m \in\{15, 50, 100, 200, 500\}$.
}
\label{fig:fp_iters_const}
\end{figure}

\begin{figure*}[h]
\vspace{1in}
\begin{center}
\includegraphics[width=0.5\linewidth]{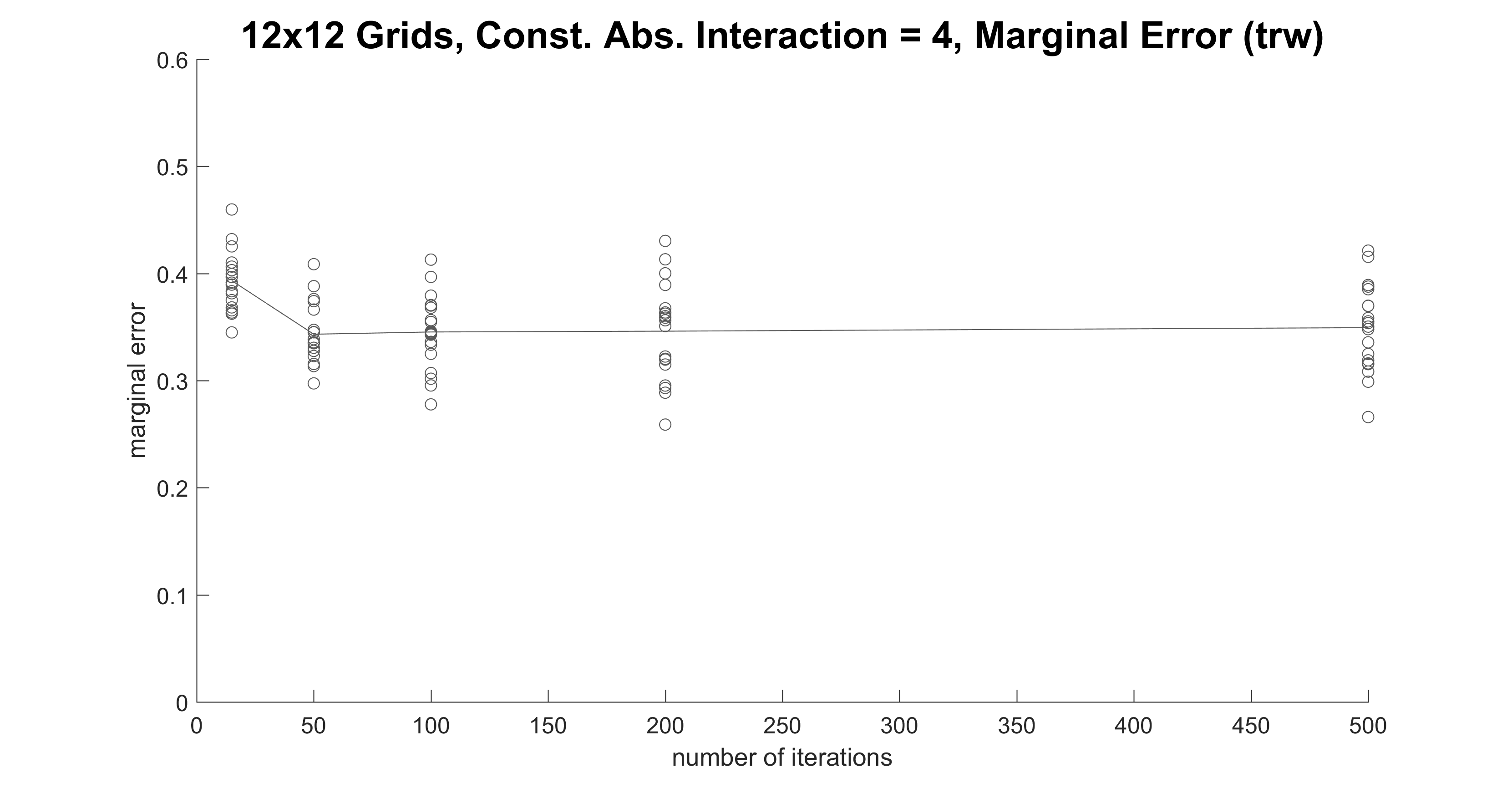}\includegraphics[width=0.5\linewidth]{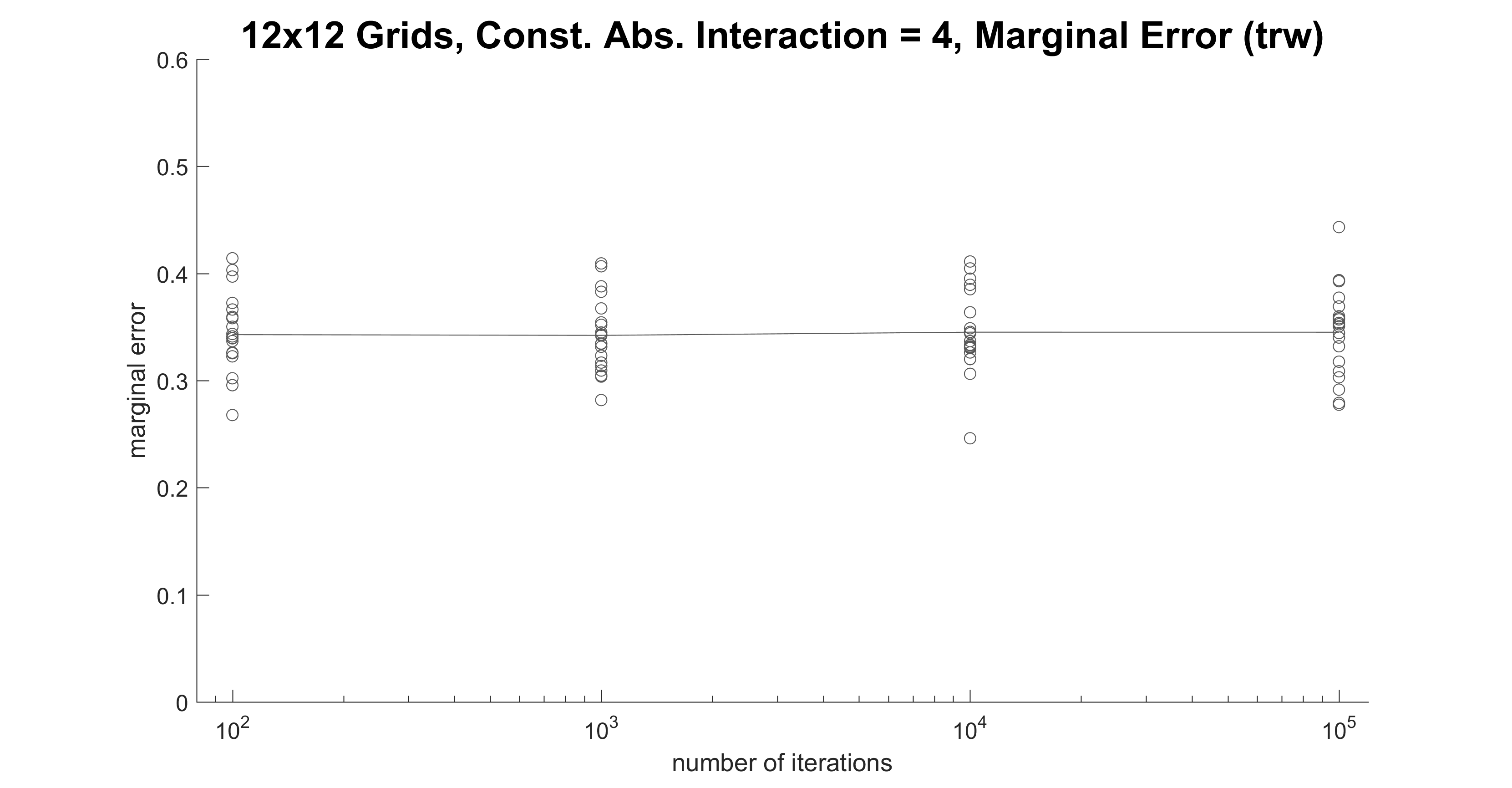}
\includegraphics[width=0.5\linewidth]{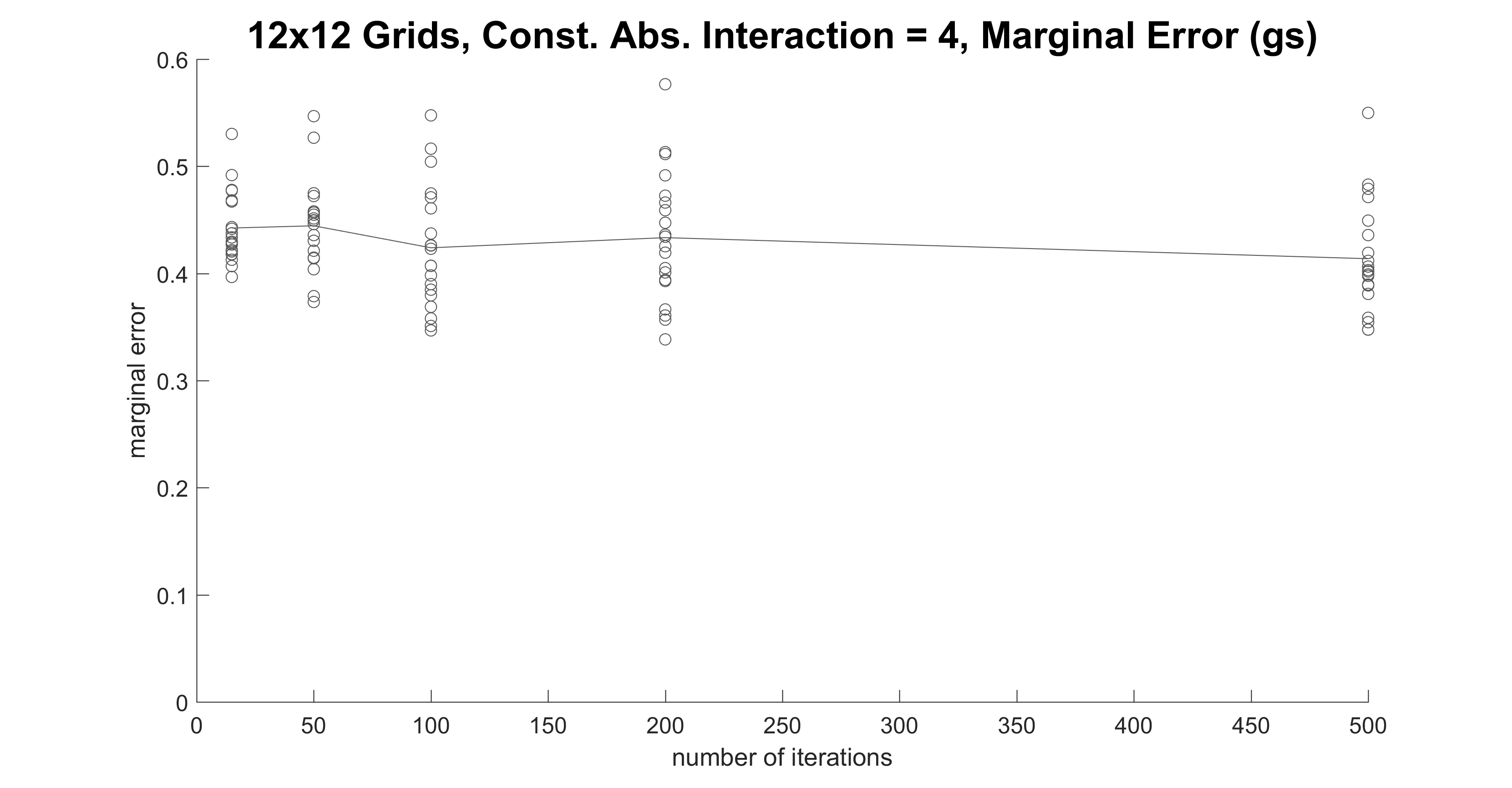}\includegraphics[width=0.5\linewidth]{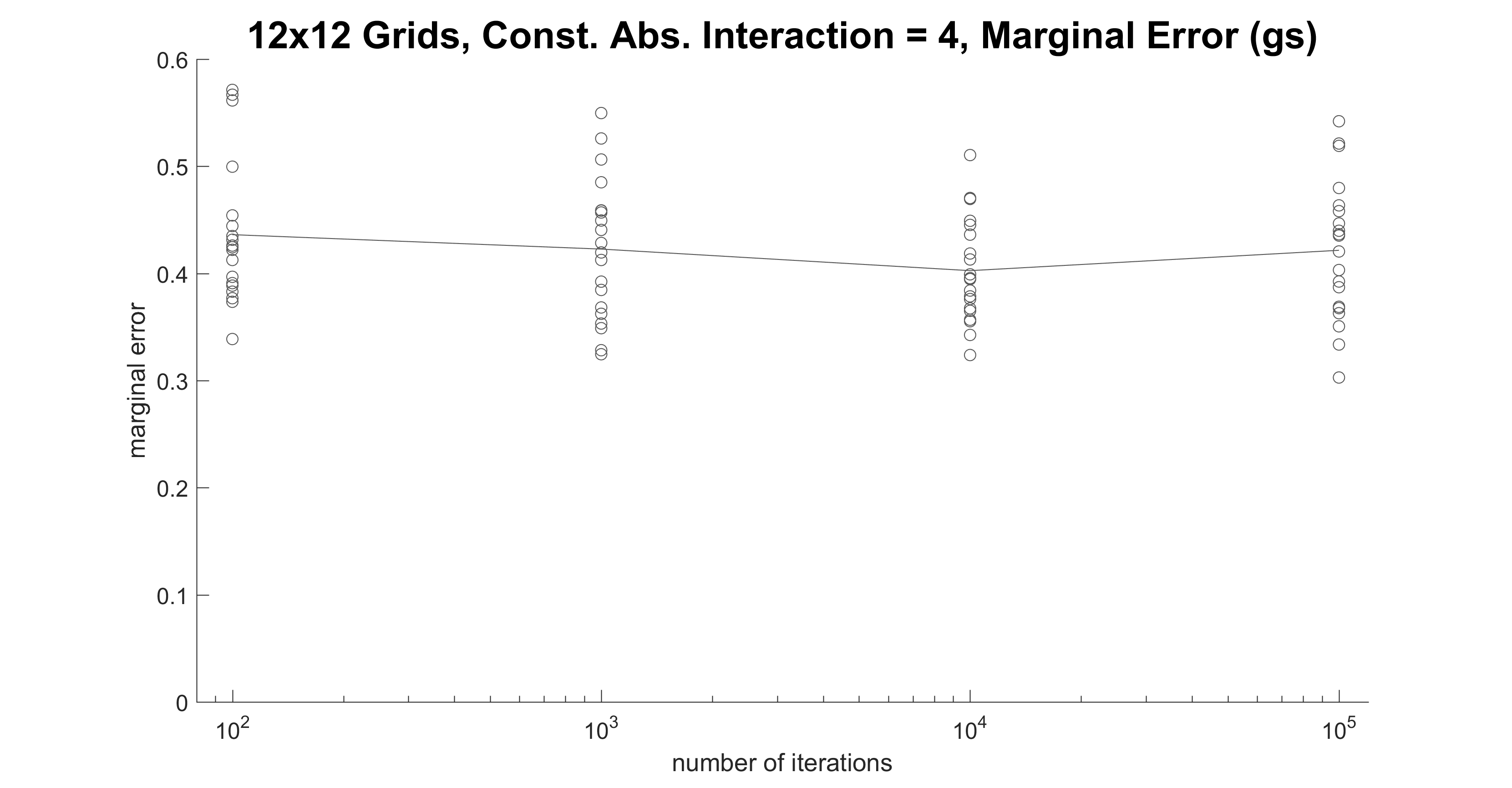}
\end{center}
\caption{{\bf Evaluation on Ising Models with 12x12 Grids, Uniform Interaction
  Magnitude (w= 4), and Varied Probability of Attractive Interactions:
  Marginal Error of trw and gs by Number of Iterations.}
This plot shows the marginal error of the estimates obtained by the trw (top plots) and gs (bottom plots) algorithms, for
different numbers of iterations.
The axes are the same as in Fig. ~\ref{fig:fp_iters_const}. Note that the right plots use a logarithmic scale for
the number of iterations.
The number of iterations used in the left plots were $m \in\{15, 50, 100, 200, 500\}$, and $m \in\{10^2, 10^3, 10^4, 10^5\}$ for the right plots.
The marginal error of each run (circles) and the average marginal error (line) were found using the 
same procedure as in Fig. ~\ref{fig:fp_iters_const}.
}
\label{fig:trw_gs_iters_const}
\end{figure*}

\paragraph{Results on the effect of the number of iterations.}
Finally, Fig.~\ref{fig:fp_iters_const} shows the marginal error of the estimates obtained by fp (ce),
for different numbers of iterations.
Increasing the number of iterations from $m=15$ to $m=50$ (and greater) only yields
minimal improvement in marginal error. 
In addition, each run of fp results in pretty consistent marginal errors at each iteration level.
Based on this, it appears that fp converges to an estimate in a fairly low number of iterations,
and does so consistently. Compare this with Fig.~\ref{fig:trw_gs_iters_const}, which shows
two similar plots each for the trw and gs algorithms. Though trw results in a comparable average
marginal error, the marginal error of each run varies more than in fp. Increasing the
number of iterations for trw does not decrease this variance, either. The same behavior occurs
with gs, though its average marginal error is a bit higher than trw and fp.

\begin{figure}[h]
\vspace{1in}
\begin{center}
\includegraphics[width=1.0\linewidth]{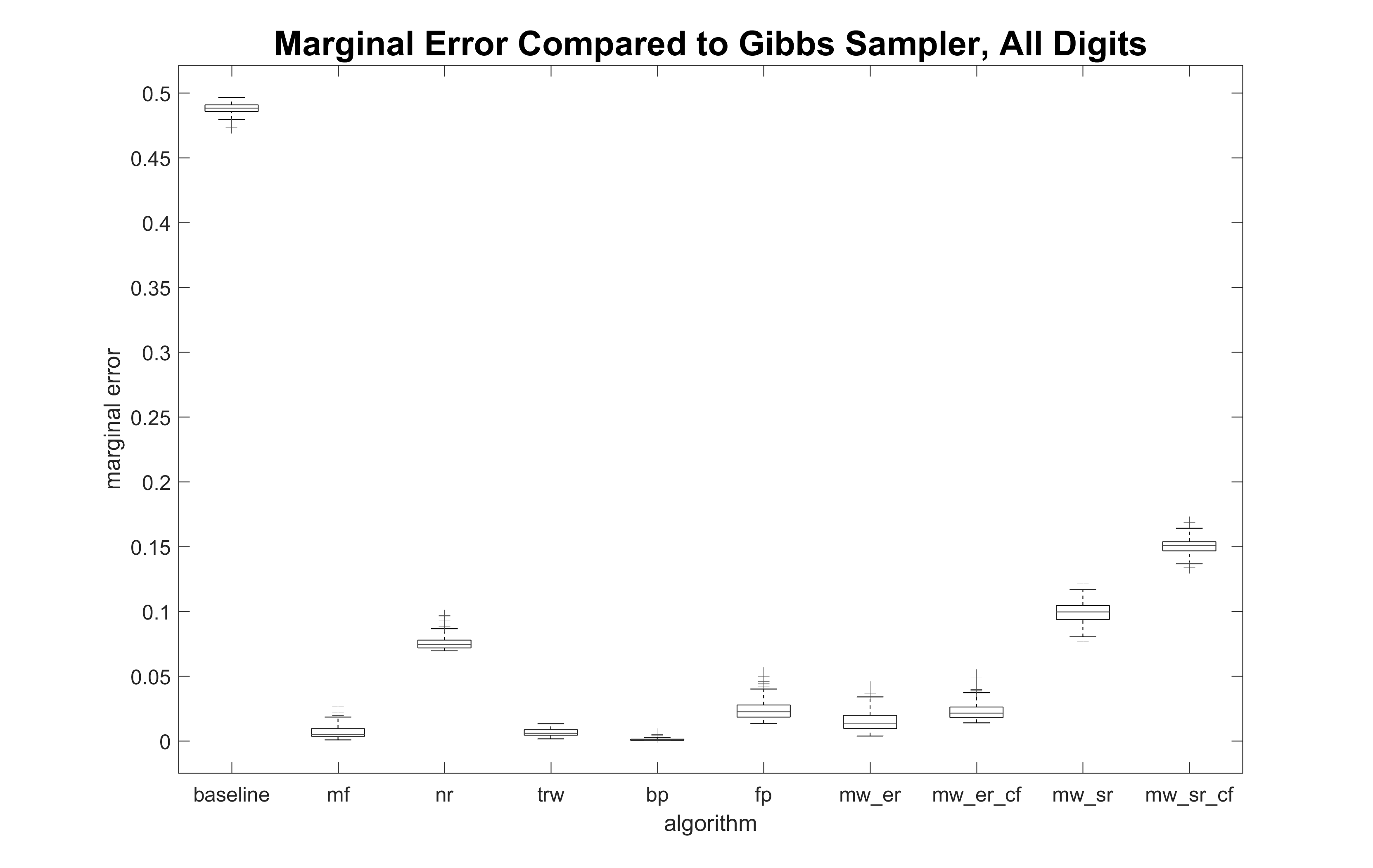}
\end{center}
\caption{{\bf Evaluation on Ising Models Derived from MNIST Images,
    28x28 Grids.}
This box plot compares the marginal errors of all algorithms with respect to gs. Because each
algorithm was run on 100 image samples, each box 
consists of 100 average (over $28^2 = 784$ variables) marginal errors. 
Data points that are more than 1.5 times the interquartile range away from the median 
are considered outliers, and are drawn as short lines. The ``whiskers'' of the boxes are
drawn as long lines.
}
\label{fig:mnist}
\end{figure}

\begin{figure}[h]
\vspace{1in}
\begin{center}
\includegraphics[width=1.0\linewidth]{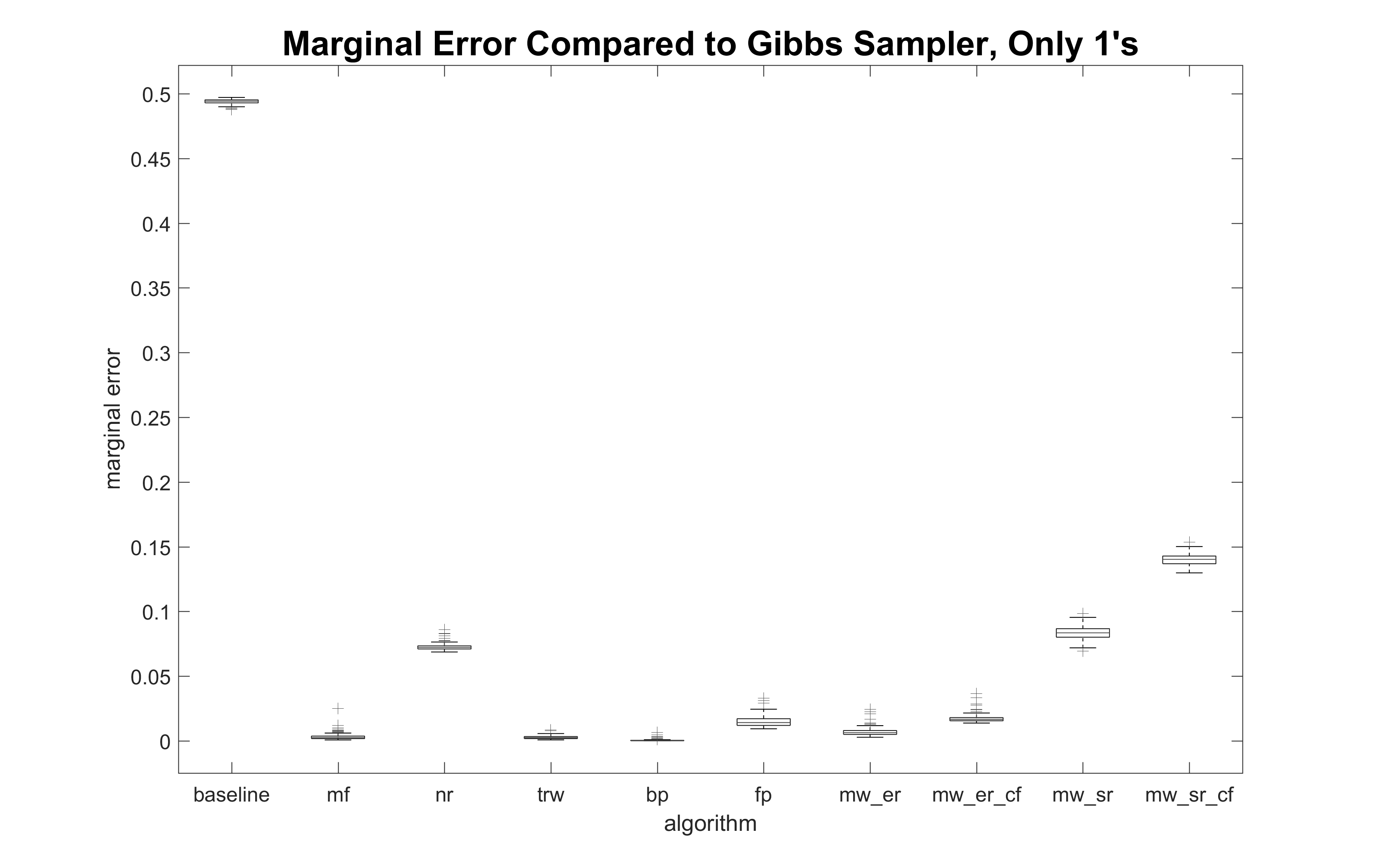}
\end{center}
\caption{{\bf Evaluation on Ising Models Derived from MNIST Images of
    Handwritten Digit "1'' Only, 28x28 Grids.}
This box plot compares the marginal errors of all algorithms with respect to gs. Because each
algorithm was run on 100 image samples, each box 
consists of 100 average (over $28^2 = 784$ variables) marginal errors. 
Data points that are more than 1.5 times the interquartile range away from the median 
are considered outliers, and are drawn as short lines. The ``whiskers'' of the boxes are
drawn as long lines.
}
\label{fig:mnist_1}
\end{figure}

\subsection{Experimental Design: MNIST-based Ising Models}

We also evaluated the various algorithms on Ising models for more
realistic settings. We use images of handwritten digits from the
popular MNIST dataset to build Ising models for soft de-noising. Note
that the interest here is not classification nor MAP estimation, but belief inference: using the
individual marginal probabilities as confidence measure on the
individual pixel values of the de-noised image. The images consist of 28x28
pixel images, so our Ising models are 28x28 simple planar
grid graphs, as in the synthetic experiments. The grayscale pixel
values in the original MNIST images
are converted to black ($+1$) and white ($-1$) values, using a threshold of $0.5$. 
Rather than randomly setting edge weights, we compute the average product between neighboring pixels, taken
across all training images, then use that average product as the edge-weight between those neighboring pixels. 
That is, denote by $I_l$ the matrix representation
of the $l$-th image in the MNIST training dataset, and denote by $m$ the
number of images in that training dataset. We set the weight $w_{(i,j),(i,j+1)}$
of the edge between
nodes/pixels $(i, j)$ and $(i, j+1)$, for example, 
\[ w_{(i,j),(i,j+1)} \propto \frac{1}{m} \sum_{l=1}^m I_l(i,j+1) \, I_l(i,j)\]
values.
and the prior bias $b_{(i,j)}$ for node/pixel $(i,j)$, for example, as
\[ b_{(i,j)} \propto \frac{1}{m} \sum_{l=1}^m I_l(i,j) \; , \]
and the normalization factor is such that $\max_{(i,j)} |b_{(i,j)}| = 1$.
We select $100$ images uniformly at random from the MNIST test
dataset, and apply the thresholding described above to turn them into a
BW images. We add $5\%$ noise to each of the resulting BW images by
"flipping'' each pixel value independently with probability $p=0.05$.
Hence, we have a different Ising model for each image: the edge
weights are all the same, but the biases differ depending on the
specific value of the test image. That is, if $I$ denotes the matrix
representation of the noisy BW test image, then the Gibbs potential of
the Ising model becomes 
\( \Psi_I(x) \equiv \sum_{((i,j),(r,s)) \in E} \widetilde{w}_{(i,j),(r,s)} \, x_{(i,j)}
\, x_{(r,s)} + \sum{(i,j)} \widetilde{b}_{(i,j)}(I(i,j)) \, x_{(i,j)}
\, , \)
where $\widetilde{w}_{(i,j),(r,s)} \propto w_{(i,j),(r,s)}$ and
$\widetilde{b}_{(i,j)}(I(i,j)) \propto b_{(i,j)} + \frac12 I(i,j)
\ln\frac{1-p}{p}$, and the normalization factor is such that
$\max_{(i,j)} |\widetilde{b}_{(i,j)}(I(i,j))| = 1$ (for consistency
with the node biases of Ising models in the synthetic experiments).

We ran the exact same algorithms on the constructed Ising models as in the synthetic experiments, though we
used a slightly different 
number of (max) iterations: $100$ for fp (all variants); $10^4$ for bp, nr, and mw (all
variants); $10^5$ for trw and gs; and $10^6$ for mf. 

\subsection{Experimental Results: MNIST-based Ising Models}

Fig. ~\ref{fig:mnist} shows our results of the experimental evaluation of the algorithms described in earlier sections (with the exception of gs).
Like in the synthetic experiments, we perform hypothesis testing using paired z-tests on the individual differences, each with p-value 0.05.
It is important to note that statements in this section comparing algorithm performance are always made with respect to gs. That is, the better performing 
algorithms here are actually producing output most similar to gs.

The computed edge-weights for the Ising models derived from MNIST images actually consist of all positive values, so this experiment 
can be thought of as analogous to the ``attractive'' case in the synthetic experiments. However, as evidenced in Fig. ~\ref{fig:mnist}, the MNIST-derived
models appear to be much ``easier'' than the synthetic models, since every algorithm performs much better than baseline. For example, mf 
has very low marginal error in this case, even though in the synthetic experiments it was often indistinguishable from bl. In order, the best
performing algorithms are 1) bp, 2) trw, 3) mf, 4) mw\_er, 5) mw\_er\_cf, 6) fp, 7) nr, 8) mw\_sr, 9) mw\_sr\_cf, and 10) bl. 

Fig. ~\ref{fig:mnist} shows our results for the case of Ising models
for the Handwritten Digit "$1$'' only.  That is, the edge weights were
computed using only images in the MNIST dataset with a training label
of the "1" digit. Likewise, the observed image samples came only from images in
the test data set with a label of the "1" digit. The figure shows that
while the various algorithms have the same relative order to each
other when run on "1's" vs "all" digits, the range of average marginal errors they achieved tightened.


\section{Future Work and New Opportunities}
\label{sec:fw}

It would be nice to have a better understanding of the exact
relationship between the \emph{true} joint distribution of the MRF and
the equilibrium points of the induced graphical potential game. For
example, it is known that no-external-regret-based algorithms like mw
converge to PSNE in ``generic'' potential games~\citep{Kleinberg:2009:MUO:1536414.1536487}, such as the
MRF-induced game. In fact, we observe such consistent convergence to
PSNE by mw in our experiments, which is unlike the behavior we
observed for fp (ce). But convergence to PSNE means that we are essentially approximating the
whole distribution with a single joint assignment (i.e., a point
mass). Yet, mw can outperform state-of-the-art algorithms like trw, 
particularly on ``hard'' instances, despite yielding such
extremely coarse approximations. In addition, best-response
dynamics in the MRF-induced game converges to a PSNE and
is equivalent to the method of \emph{iterated conditional modes (ICM)}~\citep{besag86} in PGMs, which
converges to a locally optimal joint assignment of the original
MRF. We did not include ICM in our experiments because it is generally
considered inferior to other methods. The results of mw suggests we
might want to also evaluate ICM for hard instances and compare its
output and performance to that of mw. One interesting question is
whether mw often finds better local minima that ICM, or whether ICM is
equally effective, in those hard cases. As another example of how the
proposed study would be useful, it might give us a better idea as to whether one can think of a Gibbs sampler, or other Monte-Carlo sampling algorithms, as providing solutions to equilibrium problems of certain quality.


Here we establish a connection between mf and MSNE. Despite the fact
that mf often provides poor approximations, even worst than baseline
in many cases, it would still be theoretically interesting to study
the relationship between the output of mf and that of algorithms that compute
approximate MSNE in loopy graphical games, such as {\bf NashProp}~\citep{ortizandkearns03}.

The focus of the experimental evaluation in this paper was testing our proposed, game-theoretically-inspired algorithms for belief inference with standard algorithms in the literature of probabilistic graphical models with relatively ``simple'' implementations (e.g., do not require calls to software packages or the implementation of complex optimizations). An empirical study involving such algorithms with considerably more complex implementations must have a precise experimental methodology and design that accounts for not only the complexity of implementation, but also a fair comparison that achieves the right balance between measures of solution quality and running times. We leave such evaluations for future work because of the level of complexity required to carry them out correctly.

The work in this paper just ``scratches the surface'' in terms of the synergy between equilibirum computation in game theory and belief inference in probabilistic graphical models. We state and discuss several immediate theoretical, algorithmic, and computational implications, but many more may be possible. An even broader and more thorough literature review than the one provided in this manuscript is necessary to fully exploit this connection. Thus, many opportunities for novel contributions remain available in either direction.

\section{Contributions and Concluding Remarks}
\label{sec:cont}

We provide general formulations of the problem of inference in MRFs as
equilibrium computation in graphical potential games. We provide
connections, particularly to variational inference approaches, with immediate algorithmic, computational, and theoretical
implications to belief inference in probabilistic graphical models
that follow immediately from the game-theory literature to various
related problems. We provide two approaches for approximate belief
inference: a local and a global approach. We experimentally evaluate
the effectiveness of the proposed algorithms in the context of Ising
models with grid graphs, and provide a characterization of their
computational effectiveness based on common measures used to
characterize classes of Ising models (e.g., mixed and attractive
models with different relative levels of magnitude between the edge
weights and node bias values). We also empirically evaluate
effectiveness using a slightly different approach in which we keep the
edge-weight magnitude constant but vary the ``sign probability.'' We
show how most methods are often not much better than a simple baseline
(i.e., estimate that the marginal probabilities are all equal to
$0.5$) in that class of Ising models. Our results suggest that the
proposed class of Ising models does indeed lead to harder instances
than the popular models used for empirical evaluation in the same
context of Ising models. We empirically show that our proposed method
based on a global approach
is
best, beating even TRW within that class, and shinning in a class of Ising models with
constant, ``highly attractive'' edge-weights, in which it is often better than
all other alternatives we evaluated. 
Note that TRW is generally considered
state-of-the-art. We propose such class of Ising models for future
evaluations because our experimental results suggest that instances
from that class are often the hardest. While our more local approach
is
not as effective as our global approach or TRW, in fairness, almost all of the
alternatives were no better than a simple baseline: estimate the
marginal probability to be $0.5$.

Some reviewers have expressed the view that our general
equilibrium-based approach to approximate inference is ``limited to
\emph{locally} optimal solutions to inference problems.'' We would
like to point that almost all approaches to approximate inference based on variational approximation
employed in practice, including simple methods such as mean field and state-of-the-art
methods such as TRW, suffer from exactly the same limitations.



In closing, our hope is that the work we present in this manuscript will start a conversation on the synergy between equilibrium
computation and belief inference. We believe our work and results
establish sufficient precedent for research in the direction of
formulating probabilistic inference problems as problems of equilibrium
computation. We believe this research direction is scientifically intriguing
and potentially fruitful for mathematical,
algorithmic, and computational game-theory, as well as for probabilistic graphical
models.

\bibliographystyle{plainnat} 
\bibliography{games}

\appendix
\section{Experimental Results and Discussion for 8x8 Grids}
\label{app:8x8}

Fig.~\ref{fig:std_eval} summarizes our results for the most common
classes of Ising models considered in the experimental evaluation of
approximation algorithms and heuristic for belief inference in the
literature as described above. 
We perform hypothesis testing for the result in these classes of Ising
models using paired z-tests on the individual (i.e., not joint)
differences, each with p-value $0.05$. Hence, all the statements are
statistically significant with respect to such hypotesis tests.
 Note that there is no globally best approximation technique overall
 for these classes.
\paragraph{``Mixed'' case (Left plot, Fig.~\ref{fig:std_eval}).} 
Clearly, gs is 
 best for all $w$ in this case. Among the other approximation
 algorithms, we observe the following
\begin{enumerate}
\item fp (ce) is best and better than bp for $w=4$, indistinguishable from bp
  for $w=3$, and worst than bp for $w=2$ where bp is best.
\item fp (ce) is consistently better than trw.
\item trw is worst than bp for $w<4$,
 but better than bp for $w=4$.
\item All methods, except for mf and nr, are consistently better than
  bl; mf and nr are consistently worst than bl, except for $w=2$ where mf
  is indistinguishable from bl.
\item mf and nr are indistinguishable, except for $w=4$ where nr is
  better than mf.
\end{enumerate}
\paragraph{``Attractive'' case (Right plot, Fig.~\ref{fig:std_eval}).}
In this case, there is no clear overall best. We also observe the
following.
\begin{enumerate}
\item trw is best among all methods except for $w=2$ where gs is best,
  and trw is second best.
\item fp (ce) is better than all other methods, except trw, and gs for
  $w=2$; and bp for $w < 4$ where fp (ce) is indistinguishable from bp.
\item mf, nr, bp, and gs are consistently indistinguishable
  from bl, and from each other; except for $w=2$ where gs is best, of course. 
\end{enumerate}

Fig.~\ref{fig:const_w} summarizes our experimental results for a class
of Ising models which appears to lead to ``harder'' Ising-model
instances.~\footnote{Such class of models follows from our general
  experience with similar models. We find that instantiating Ising
  model parameters using densities over edge-weights tended to yield
  to relatively easier models than the ones we obtain by fixing the
  magnitude of the edge-weights and varying the probability of their
  sign, independently for each edge.} 
We perform hypothesis testing for the result in these classes of Ising
models using two approaches depending on $w$. For $w=4$, where we draw
$50$ models as samples for each $q$, we use appropriately modified paired z-tests on the individual (i.e., not joint)
differences, each with p-value $0.05$. We modify the calculation of
the variances resulting from the average over the samples computed for
each $q$. We do so because the distributional properties of the empirical
mean/average for each $q$ may differ. 
For $w < 4$, where we only draw $5$ models as samples for each $q$, we
use bootstrapped-based, individual, paired hypothesis-testing over
each pair of aggregate differences between the methods for each of
those values of $w$; we use $100$ bootstrap samples, and p-value $0.05$
All the statements are
statistically significant with respect to such hypotesis tests.
\paragraph{Aggregate results (Left plot, Fig.~\ref{fig:const_w}).}
The left-hand plot in Fig.~\ref{fig:const_w} shows the aggregate results for
this case. There is no clear
overall best over all $q$. We also observe the
following.
\begin{enumerate}
\item fp (ce) is best for $w=4$ and $w=3$, while being second best
  to gs for $w=2$; and, for $w=2.5$, tied for best with gs
  (i.e., indistinguishable from gs).
\item trw is consistently better than bp, mf, bl, and nr, except for
  $w=2.5$ where trw is indistinguishable from bp; trw is consistently
  worst than fp (ce)
\item Only mf is worst than bl for $w\in\{2,4\}$; mf is
  indistinguishable from bl for $w\in\{2.5,3\}$; also, bp and nr are
  indistinguishable from bl, except for $w=2.5$, where bp is better
  than bl.
\item bp is better than mf except for $w=2$; bp is indistinguishable from nr for $w\in\{2,3\}$, but bp is
  better than nr for $w\in\{2.5,4\}$.
\item gs is consistently better than mf and nr; indistinguishable from trw, except, of course, for $w=2$ where gs is tied for
  best with fp (ce).
\end{enumerate}

\paragraph{Results for constant edge-weight magnitude $w=4$ as a
  function of probability of attractive interaction $q$ (Right plot, Fig.~\ref{fig:const_w}).}
The right-hand plot in Fig.~\ref{fig:const_w} shows finer-grain results
for this case.
The results suggest that in fact
such instances of Ising models tend to be harder in the sense that even
state-of-the-art algorithms such as TRW are no better than the simple
baseline estimation, in which $\widehat{p}_i=0.5$ for all
nodes/variables $i$, for less than half of the full range of values of
the sign probability $q$ (i.e., for $q \in
\{0.1,0.4,0.5,0.6,0.8,0.9\}$). In fact, the performance of TRW is
\emph{almost exactly the same} as baseline accros the range of
non-extreme values of $q$. (Note how the plot of the values for trw
and bl are essentially on top of each other for values of $q$ other than $0$ or $1$.) 
On the other hand, note how fp (ce) is consistently better than bl
across the whole range of values for $q$. In fact, fp (ce) is always in the set of (statistically)
best performers for all $q$: i.e., the single best for $q=1.0$, and indistinguishable from trw for $q=0.0$;
gs and bp for $q\in\{0.1,0.5,0.6,0.7\}$; nr, gs, and bp for $q=0.2$;
gs for $q\in\{0.3,0.4,0.9\}$; and gs and mf for $q=0.8$.  The proposed fp (ce) is also best at both extremes,
while trw is only best when all weights are negative. Almost all the
methods other than fp (ce) are no better, and often worst, than bl, except for bp and
trw for $q=0.0$; trw for $q\in\{0.2,0.7,1.0\}$; trw and gs for
$q=0.3$; and gs
for $q=0.8$.

\end{document}